\documentclass[lettersize,journal]{IEEEtran}
\usepackage{caption}
\usepackage{subcaption}
\usepackage{amsmath,amsfonts}
\usepackage{algorithmic}
\usepackage{algorithm}
\usepackage{array}
\usepackage{textcomp}
\usepackage{stfloats}
\usepackage{gensymb}
\usepackage{url}
\usepackage{verbatim}
\usepackage{graphicx}
\usepackage{cite}
\usepackage{wrapfig}
\usepackage{amssymb}
\usepackage[dvipsnames]{xcolor}
\usepackage{amsmath}
\usepackage{makecell}
\usepackage{threeparttable}
\usepackage{hyperref}
 \usepackage{array,multirow,graphicx}
 \usepackage{float}
\hypersetup{colorlinks,linkcolor={blue},citecolor={blue},urlcolor={red}} 
\newtheorem{prop}{\textbf{Proposition}}
\newtheorem{definition}{\textbf{Definition}}
\newcolumntype{P}[1]{>{\centering\arraybackslash}p{#1}}
\newcolumntype{M}[1]{>{\centering\arraybackslash}m{#1}}

\begin{document}

\title{TraNCE: Transformative Non-linear Concept Explainer for CNNs}

\author{\IEEEauthorblockN{
Ugochukwu Ejike Akpudo,
Yongsheng Gao~\IEEEmembership{Senior Member,~IEEE,}
Jun Zhou, and
Andrew Lewis}

\IEEEauthorblockA{Integrated and Intelligent Systems, Griffith University, Australia}

ugochukwu.akpudo@griffithuni.edu.au, \{yongsheng.gao, jun.zhou, a.lewis\}@griffith.edu.au

%
}



\maketitle

\begin{abstract}
Convolutional neural networks (CNNs) have succeeded remarkably in various computer vision tasks. However, they are not intrinsically explainable. While feature-level understanding of CNNs reveals where the models looked, concept-based explainability methods provide insights into what the models saw. However, their assumption of linear reconstructability of image activations fails to capture the intricate relationships within these activations. Their Fidelity-only approach to evaluating global explanations also presents a new concern. For the first time, we address these limitations with the novel Transformative Non-linear Concept Explainer (TraNCE) for CNNs. Unlike linear reconstruction assumptions made by existing methods, TraNCE captures the intricate relationships within the activations. This study presents three original contributions to the CNN explainability literature: (i) An automatic concept discovery mechanism based on variational autoencoders (VAEs). This transformative concept discovery process enhances the identification of meaningful concepts from image activations. (ii) A visualization module that leverages the Bessel function to create a smooth transition between prototypical image pixels, revealing not only what the CNN saw but also what the CNN avoided, thereby mitigating the challenges of concept duplication as documented in previous works. (iii) A new metric, the \textit{Faith} score, integrates both Coherence and Fidelity for comprehensive evaluation of explainer faithfulness and consistency. Based on the investigations on publicly available datasets, we prove that a valid decomposition of a high-dimensional image activation should follow a non-linear reconstruction, contributing to the explainer's efficiency. We also demonstrate quantitatively that, besides accuracy, consistency is crucial for the meaningfulness of concepts and human trust.  The code is available at \url{https://github.com/daslimo/TrANCE}

\end{abstract}


\section{Introduction}
\label{sec:intro}

\IEEEPARstart{C}{onvolutional} neural networks (CNNs) are widely used in computer vision, achieving notable success in visual classification tasks~\cite{cnnreview_jun,obj_detection_review}. However, understanding them at a human level remains a major challenge in artificial intelligence (AI), raising significant concerns about their explainability, especially in promoting ethical AI~\cite {visual_answer, review_xai, juliusadebayo}. Significant initiatives aim to establish interpretability as essential for AI adoption, ensuring trust and transparency due to economic and personal impacts~\cite{Baobao, AAAITREND, gdpr}. In 2022, Stanford's AI Index reported 37 AI-related laws enacted in 127 countries, up from one in 2016~\cite{stanford_ai_impact}. A recent survey shows 68\% of respondents believe AI safety should be prioritized, especially in military and health sectors~\cite{Baobao}. These highlight the importance of interpretability (alongside accuracy) for responsible use of AI~\cite{graph_explain, ai_impact, abejide}. 

\begin{figure}[!t]
\centering
\includegraphics[width=1\linewidth]{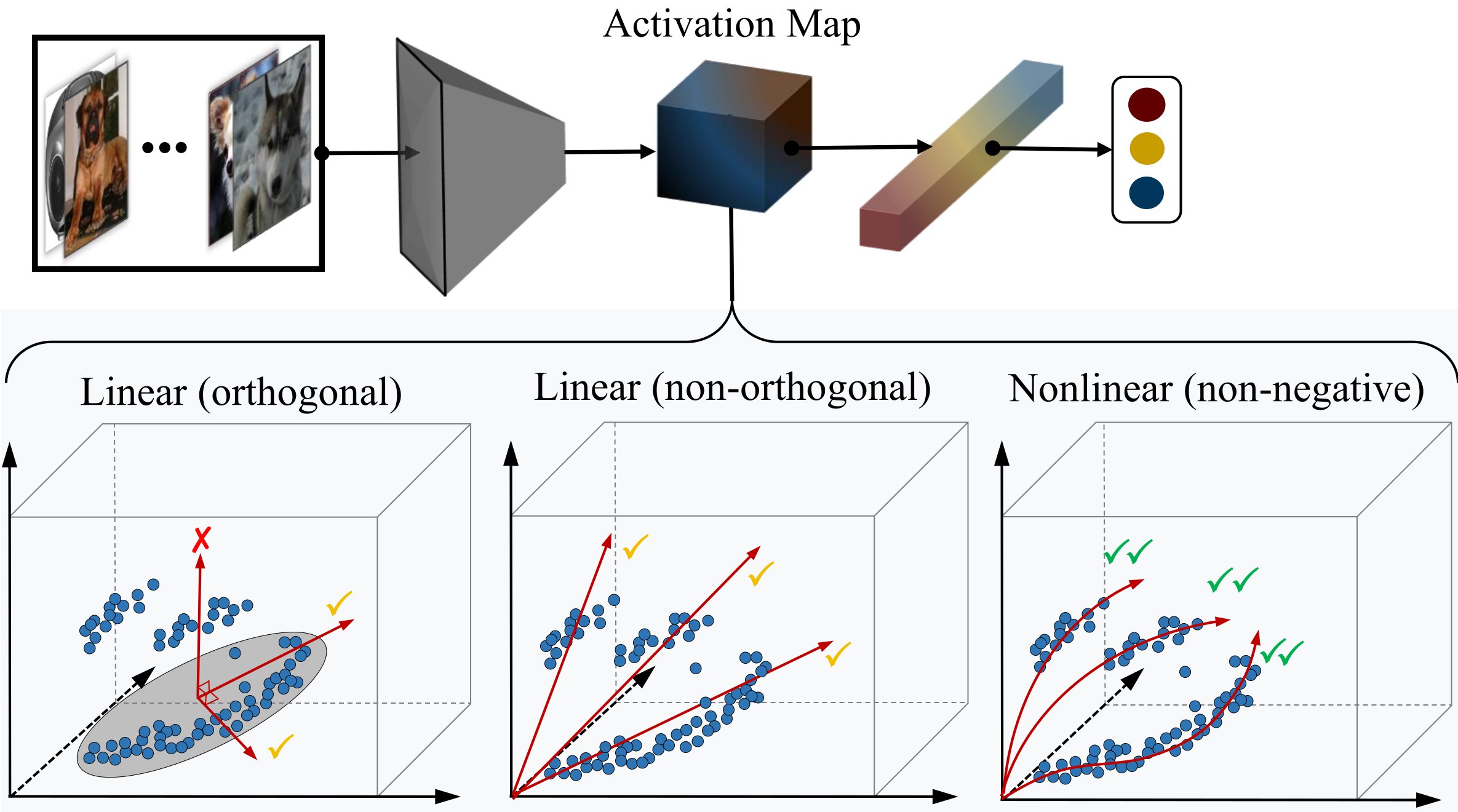}
\caption{The rationale for non-linear reconstructability of image activations produced by a CNN in the latent space. An orthogonal linear decomposition may produce invalid latent projections (\textcolor{red}{×} mark) while the non-orthogonal linear decomposition may produce acceptable latent projections as non-negative projections (\textcolor{Goldenrod}{\checkmark} mark). In contrast, a non-linear decomposition produces valid non-negative latent projections (\textcolor{ForestGreen}{\checkmark\checkmark} mark).}
\label{graphical1}
\end{figure}

The conventional feature-level understanding of CNNs raises a notable concern: their explanations are often qualitative and therefore susceptible to bias; they only provide insight into \textit{where} the model looked~\cite{gradcam_blackbox, review_xai, zohaib, juliusadebayo, Baobao, channel_prune_saliency}. Concept-based explanations~\cite{tcav, ice, ace, craft, acevideo, zebra, glance, hint, pipnet, labo} offer both qualitative explanations and quantitative checkpoints that support better human understanding, providing insight into \textit{what} the model saw (as concepts)~\cite{tcav, overlookedfactors}. Despite the early success of pioneering concept-based explanation methods~\cite{tcav, glance, hint, pipnet, labo}, they rely on pre-defined concepts~\cite{overlookedfactors, craft}. In contrast, integrating dimensionality reduction methods with CNNs enables automatic concept discovery without explicit human supervision~\cite{ace, ice, craft}. However, linear reducers are deterministic, assuming linear reconstructability of inputs and failing to capture intricate relationships within image activations. As shown in Fig.~\ref{graphical1}, an orthogonal linear reducer can produce invalid projections (marked by \textcolor{red}{×}). While a non-orthogonal linear reducer might yield acceptable non-negative projections (marked by \textcolor{Goldenrod}{\checkmark}), a non-linear reducer better fits the data structure, producing valid non-negative projections (marked by \textcolor{ForestGreen}{\checkmark\checkmark}). Empirically, the lack of adequate quantitative checkpoints for faithfulness evaluation of concept explanations remains another significant concern. Empirically, the lack of adequate quantitative checkpoints for evaluating the faithfulness of concept explanations remains a significant concern~\cite{akpudoicem2024}. Existing methods rely solely on Fidelity--the level of agreement of the concept predictions with the input predictions--for global faithfulness evaluation~\cite{ice, akpudoicem2024,why_trust_you}, which unfortunately, is limited as it excludes Coherence--the measure of (in)consistency of concept explanations with CNN predictions.

This article proposes a novel concept-based explanation framework for CNNs that automatically discovers accurate and consistent concepts, capturing the intricate relationships within image activations for enhanced local and global explanations to address these issues. The proposed TraNCE exploits the robustness of a variable autoencoder (VAE) for non-linear decomposition of the CNN's high dimensional feature map for automatic concept discovery. It introduces a quantitative measure of the consistency of the concept explanations. On the other hand, it offers improved trustworthiness evaluation of the explanations as shown in our experiments on fine-grained datasets. The contributions of this paper are summarized as follows:

\begin{itemize}
\item We propose a novel Transformative Non-linear Concept Explainer (TraNCE), an unsupervised concept-based explanation framework for CNNs. TraNCE employs a non-linear reducer, VAE for automatic concept discovery that better captures the intricate relationships within the image activations, thereby providing enhanced local and global CNN explanations.
\item We harness the transformative architecture of the VAE's decoder and introduce Coherence as a quantitative measure of concept consistency. Combined with Fidelity, it forms the comprehensive \textit{Faith} score, enhancing the evaluation of global explanation faithfulness. We demonstrate the importance of evaluating model faithfulness beyond the traditional use of Fidelity scores. 
\item TraNCE outperforms baseline methods, showcasing efficacy in our experiments on Fine-Grained Visual Categorization (FGVC) tasks. We mitigate the problems of returning duplicate and ambiguous concepts that are not human-understandable by introducing a more human-understandable visualisation module that reveals precisely what a CNN saw.

\item For the first time, we investigate the impact of image transformations on explainer sanity i.e. the degree to which the explanations provided by the CNN Explainer are logical and in line with expectations. We conducted sanity checks on the TraNCE framework, highlighting key concerns and suggesting directions for future work.
\end{itemize}

The rest of the paper is organised as follows: Section \ref{sec:motivation} presents the motivation of our study and a review of related works, Section~\ref{background} presents the preliminaries for our study, Section~\ref{proposed_mtd} presents the proposed TraNCE explanation framework and its constituents, Section~\ref{experiments} presents the experiments, Section~\ref{discussion} discusses TraNCE's impact, potentials and limitations, and Section~\ref{conclusion} concludes the paper.

\section{Motivation and Related Work}
\label{sec:motivation}

The increasing demand for transparency has emphasized the significance of Explainable AI (XAI) \cite{graph_explain, abejide, stanford_ai_impact, Baobao, ai_impact}. The complexity of CNN models, the trade-off between accuracy and interpretability, and the lack of standardized interpretability frameworks impede the advancement of XAI. This necessitates thorough and understandable explanations to humans \cite{ace, overlookedfactors, cam_metrics, complete_ex}. Different perspectives have been proposed including feature visualization, concept activation vectors, and attribution/saliency for CNN understanding.

A recent discovery~\cite{abejide} reveals that attribution/saliency methods ~\cite{gradcam, gradcanplusplus, fullgrad, hirescam, ablationcam, eigencam, layercam, gradcam_blackbox, pylon, zohaib} lead to selective, conceptual, and confirmation biases. They fundamentally provide information on \textit{where} the model looked, not \textit{what} it saw, therefore cannot establish causality. These methods have a common limitation: large CNN models may produce noisy gradients, leading to misestimates of pixel importance and similar estimates for different classes \cite{juliusadebayo}. Some of these methods also quantify and rank the contributions of individual features to a model's predictions~\cite{why_trust_you, channel_prune_saliency, pane, pylon}. However, they may overestimate the significance of irrelevant features or underestimate relevant ones. Their effectiveness varies by model type, they cannot establish causality, and they are sensitive to data perturbations.  

Concept-based explainers, which represent high-level patterns or abstract ideas within an image class, offer more intuitive and human-understandable explanations \cite{tcav,ace}. These explainers quantify the significance of a concept using Concept Activation Vectors (CAVs), which measure how much a concept (e.g., gender, race) influences a model's prediction, providing insights on \textit{what} a CNN saw \cite{ice,craft}. Methods like TCAV \cite{tcav}, HINT \cite{hint}, GLANCE \cite{glance}, PIP-NET \cite{pipnet}, and ATL-Net \cite{labo} quantify the influence of predefined concepts on CNN predictions. However, these methods are computationally expensive and require probe datasets with predefined concepts, which can impact trust when labelled concepts are unavailable \cite{ace,overlookedfactors}. The choice of probe dataset also significantly affects the quality of explanations, as concepts in the probe dataset can be more challenging to learn than the target classes \cite{overlookedfactors}. These challenges highlight the need for unsupervised concept-based explanation methods for automatic concept discovery.

State-of-the-art unsupervised concept-based explanation methods \cite{ace, ice, craft} support automatic concept discovery. However, they exhibit significant drawbacks. The ACE framework \cite{ace} causes information loss during outlier rejection and produces varying concept weights for different instances. ICE \cite{ice} and CRAFT \cite{craft} address some limitations by using non-negative matrix factorization (NMF) instead of ACE's clustering method, allowing for the quantification of information losses. However, the assumption that high-dimensional CNN feature maps can be linearly reconstructed by NMF is problematic, especially for FGVC tasks with small inter-class variance and complex, non-linear feature relationships. Additionally, their Fidelity-only approach to faithfulness evaluation presents concerns \cite{akpudoicem2024}. Therefore, employing a more robust reducer is justified \cite{ncaf}. Variational Autoencoders (VAEs) effectively capture complex, non-linear relationships within feature maps, making them ideal for intricate scenarios \cite{vae_explain, vae_continuous, vae_parameters}. Their probabilistic nature allows for flexible latent representations, enhancing data distribution modelling \cite{vae_continuous}. Additionally, VAEs' adaptability and continuous latent space improve utility for various explanation tasks.

\begin{figure*}[!tbph]
\centering
\begin{minipage}[b]{.98\linewidth}
\centering\centerline{\includegraphics[width=1\linewidth]{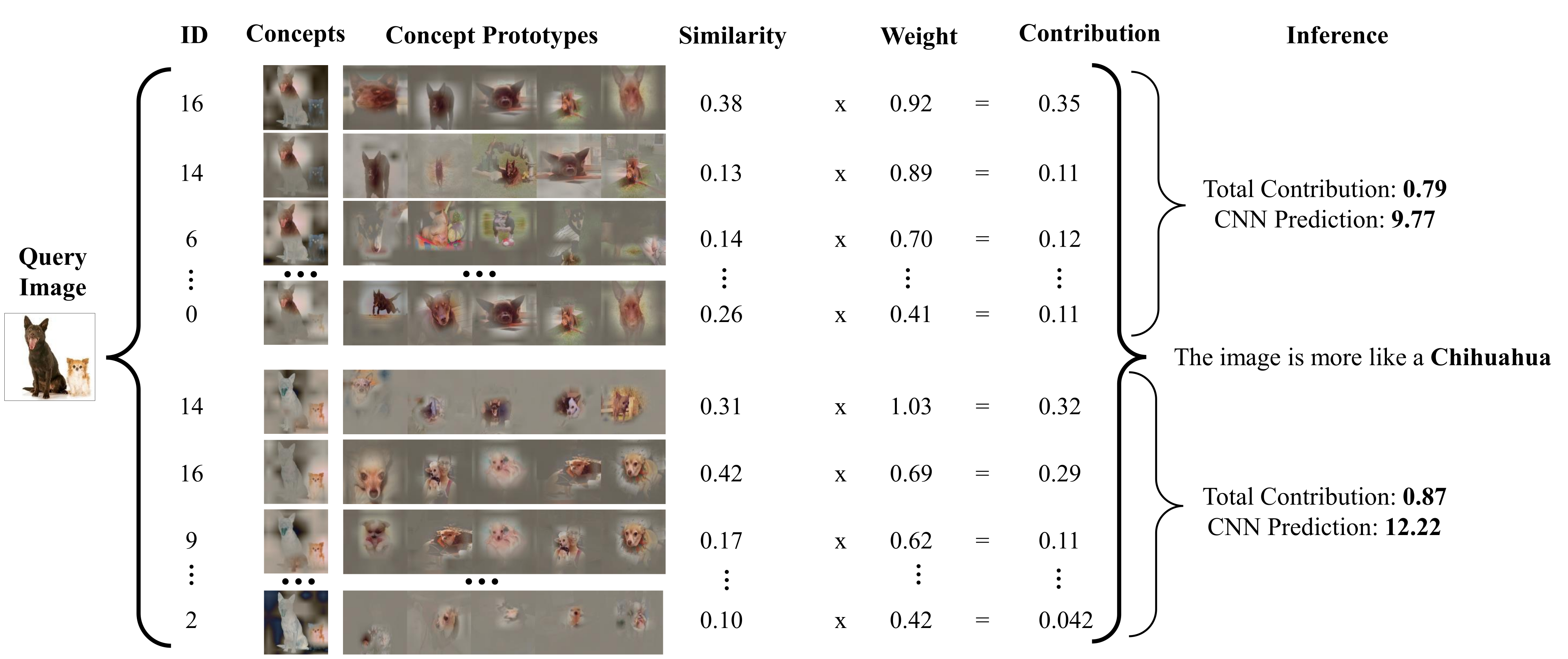}}
  \centerline{(a)}\medskip
\end{minipage}

\begin{minipage}[b]{.98\linewidth}
  \centering
  \centerline{\includegraphics[width=1\linewidth]{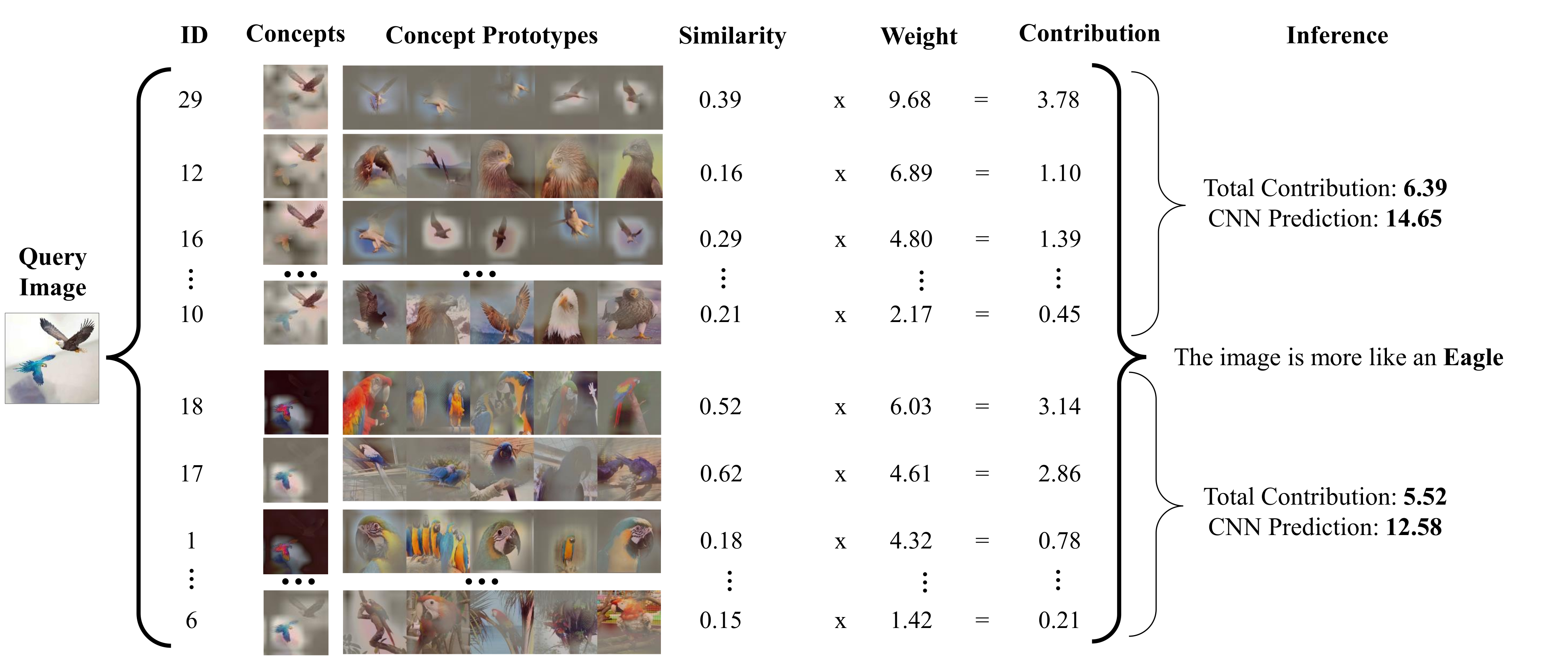}}
  \centerline{(b)}\medskip
\end{minipage}
\centering
\vspace*{-1mm}
\caption{Local explanations for (a) ResNet50, tested on an image of an Australian Kelpie and a Chihuahua, and (b) InceptionV3, tested on an image of a Macaw and an Eagle. TraNCE produces prototypes for each concept by automatically optimizing image regions from the training images that faithfully represent a concept in the target class. The total \texttt{Contribution} for each target class is the sum of the product of \texttt{Weight}s and the Similarity scores.}
\label{local_explanations}
\end{figure*}

A robust explainer, demonstrated in Fig.~\ref{local_explanations} should provide accurate, consistent, and understandable explanations, supported by quantitative insights to enhance trust \cite{ace}. This underscores the importance of diversified evaluation checkpoints in XAI~\cite{cam_metrics}. While the VAE's inverse function quantifies information losses, relying solely on Fidelity presents concerns \cite{MAPE}. Fidelity, often measured by mean absolute percentage error (MAPE) or R-squared in a few cases~\cite{why_trust_you,integrated_gradients,gradientshap}, has limitations such as issues with low-volume data, ignoring zero values, rewarding inaccurate forecasts with small values, and skewness by outliers \cite{MAPE}. Although Fidelity offers insights into concept explanation accuracy, it does not address the consistency of the concept explanation. To overcome this, we propose integrating Coherence with Fidelity to form a new metric, the \textit{Faith} score, providing a more comprehensive assessment of an explainer's faithfulness such that the pros of one mitigate the cons of the other. Inspired by TraNCE, this approach aims to improve concept discovery and faithfulness evaluation.

\section{Preliminaries}
\label{background}

\subsection{Probing into a CNN}
CNN-based FGVC tasks typically involve supervised learning scenarios, where input images consisting of pixel values $(x_1, \ldots, x_n) \in \mathcal{X}^n$ are mapped to associated labels ($y_1, \ldots, y_n) \in \mathcal{Y}^n$ \cite{resnet, wideresnet, resnext}. Consider a CNN model $f(\cdot) \equiv f: \mathcal{X} \rightarrow \mathcal{Y}$ such that $f(x)$ is the trained CNN model that produced $h_l^k(x)$ activations at the $l^{th}$ of $k$ layers ($l = 1, 2, 3, \ldots, k$). 

Practically, $f(\cdot)$ can be separated into two parts at the intermediate layer: the concept (feature) extractor $E(\cdot)$ and the classifier $C(\cdot)$. $E(x)$ produces the intermediate feature activations $\mathcal{A}_{l} \in \mathbb{R}^D$. $D = (h, w, c)$, where $h \times w$ is a single feature dimension of height $h$ and width $w$ while $c$ is the number of channels. $\mathcal{A}_{l}$ contains $c$ discriminative features with complex and non-linear relationships which can be exploited for explaining the CNN decision logic. Practically, the goal is to summarise $\mathcal{A}_{l}$ from $\mathbb{R}^D$ to $\mathbb{R}^d$ to produce $c^\prime$ discriminative features such that $d= (h, w, c^\prime)$. $d \ll D$ since $c^\prime \ll c$.

\subsection{non-linear Featuremap Decomposition}
Given the high-dimensional $\mathcal{A}_{l}$, the computation of eigenvectors is an essential step in the majority of linear decomposition techniques like the PCA~\cite{pca_variants, automate}. PCA produces orthogonal embedding with lower dimensions that are assumed to be isometric. However, such orthogonal embedding struggle to represent intricate and non-linear structures in $\mathcal{A}_{l}$. When $\mathcal{A}_{l}$ cannot be isometrically embedded as a Euclidean subspace of $\mathbb{R}^{D}$, as is the case in FGVC tasks, $\mathbb{R}^{d}$ fails to capture the intrinsic geometric structure of $\mathcal{A}_{l}$. 

Other linear decomposition methods like the NMF factorize $\mathcal{A}_{l}$ into two lower-dimensional non-orthogonal non-negative matrices by minimizing the Frobenius norm~\cite{nndsvd, ammd}. While NMF offers acceptable utility because it enforces non-negativity constraints, an advantage in image processing, its assumption that $\mathcal{A}_{l}$ has a linear structure is expensive, leading to an over-simplistic decomposition of $\mathcal{A}_{l}$. In addition to these limitations, linear reducers can be unstable in response to perturbations of $\mathcal{A}_{l}$. Particularly, the presence of stochastic elements in the initialization phase of NMF results in non-orthogonal non-negative matrices that vary at every new run. Also, the sparsity of $\mathcal{A}_{l}$ can cause serious distortions, especially when $\mathcal{A}_{l}$ cannot be linearly factorized in the subspace of $\mathbb{R}^{D}$. 

For FGVC tasks with small inter-class variance, $\mathcal{A}_{l}$ has non-linear $c$ discriminant features that are complex. Consequently, manifold learning (non-linear dimensionality reduction such as Isomaps, Locally Linear Embedding, Kernel PCA, and VAEs~\cite{dimesnion_mtds}) achieves better decomposition of $\mathcal{A}_l$ while retaining its integrity \cite{pca_variants, dimesnion_mtds} (see Fig~\ref{graphical1}). However, VAEs offer superior benefits in the following ways: their probabilistic framework allows them to adapt to diverse datasets and complex data distributions, their superior capacity for generative tasks, and their potential for interpretable representations within the latent space \cite{vae_continuous, vae_explain, vae_parameters, vae_parameters2, onceclassVAE}. 

\subsection{Bessel Functions for Concept Visualization}

Named after the German mathematician Friedrich Bessel, Bessel functions are fundamental solutions to Bessel’s differential equation, frequently encountered in problems featuring cylindrical or spherical symmetry, e.g. heat conduction, wave propagation, and electromagnetic fields \cite{bessel}. Despite their

\begin{wrapfigure}{r}{0.4\linewidth}
  \centering
   \vspace*{0mm}\includegraphics[width=1\linewidth]{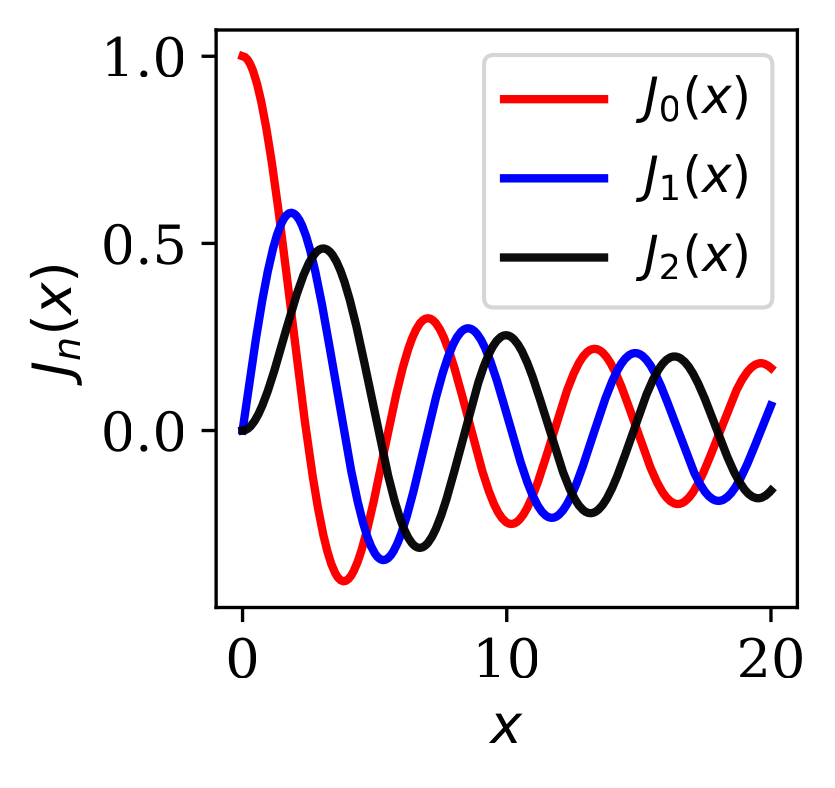}
\caption{Oscillation of Bessel function of first kind of integer order at $J_0(x), J_1(x)$, and $J_2(x)$.}
\label{bessel_plot}
\end{wrapfigure}

\noindent broad applicability, the Bessel function $J_\nu(x)$ ($\nu \subset \mathbb{R}$) of the first kind, characterized by order $n$ defined in Eq.~(\ref{besss}), serves as a valuable tool for heatmap visualization. Leveraging its inherent smoothness and orthogonality properties, it enables quantitative analysis of heatmap patterns, offering visual insights into the underlying data structure. Fig.~\ref{bessel_plot} illustrates the oscillation of $J_\nu(x)$ of the first kind of integer order.
\begin{equation}
\label{besss}
    J_\nu(x) = \sum_{r=0}^\infty \frac{(-1)^r}{r! (\nu + r)!} \left(\frac{x}{2}\right)^{\nu + 2r},
    \left\{\begin{array}{c}
    1 \text { if } x,\nu = 0 \\
    0 \text { if } \nu > 0\\
    (-1)^\nu J_\nu(x) \text { if } \nu < 0\end{array}\right.
\end{equation}
where $\nu$ is a real or complex parameter and $x$ is the independent variable i.e. the pixel values of an image.

As shown in Fig.~\ref{bessel_plot}, the horizontal axis represents the independent variable $x$, while the vertical axis represents the Bessel functions $J_n(x)$. Each curve corresponds to a different Bessel function: $J_0(x)$ (red) oscillates around zero with a single maximum, indicating high values of $J_\nu(x)$; $J_1(x)$ (blue) oscillates with one zero crossing, indicating intermediate values; and $J_2(x)$ (dark grey) oscillates with two zero crossings, indicating low values. 

Concept visualization represents discovered concepts as highlighted pixels with high activations to a target class. In Fig.~\ref{bessel_plot}, red indicates strong activations, grey indicates low activations, and blue indicates moderate activations. The sinusoidal colour mapping from the Bessel function creates smooth transitions, revealing intuitive visualizations.

\section{Proposed Method}
\label{proposed_mtd}

\subsection{Overview of TraNCE}

\begin{figure*}[t]
\centering
\includegraphics[width=0.98\textwidth]{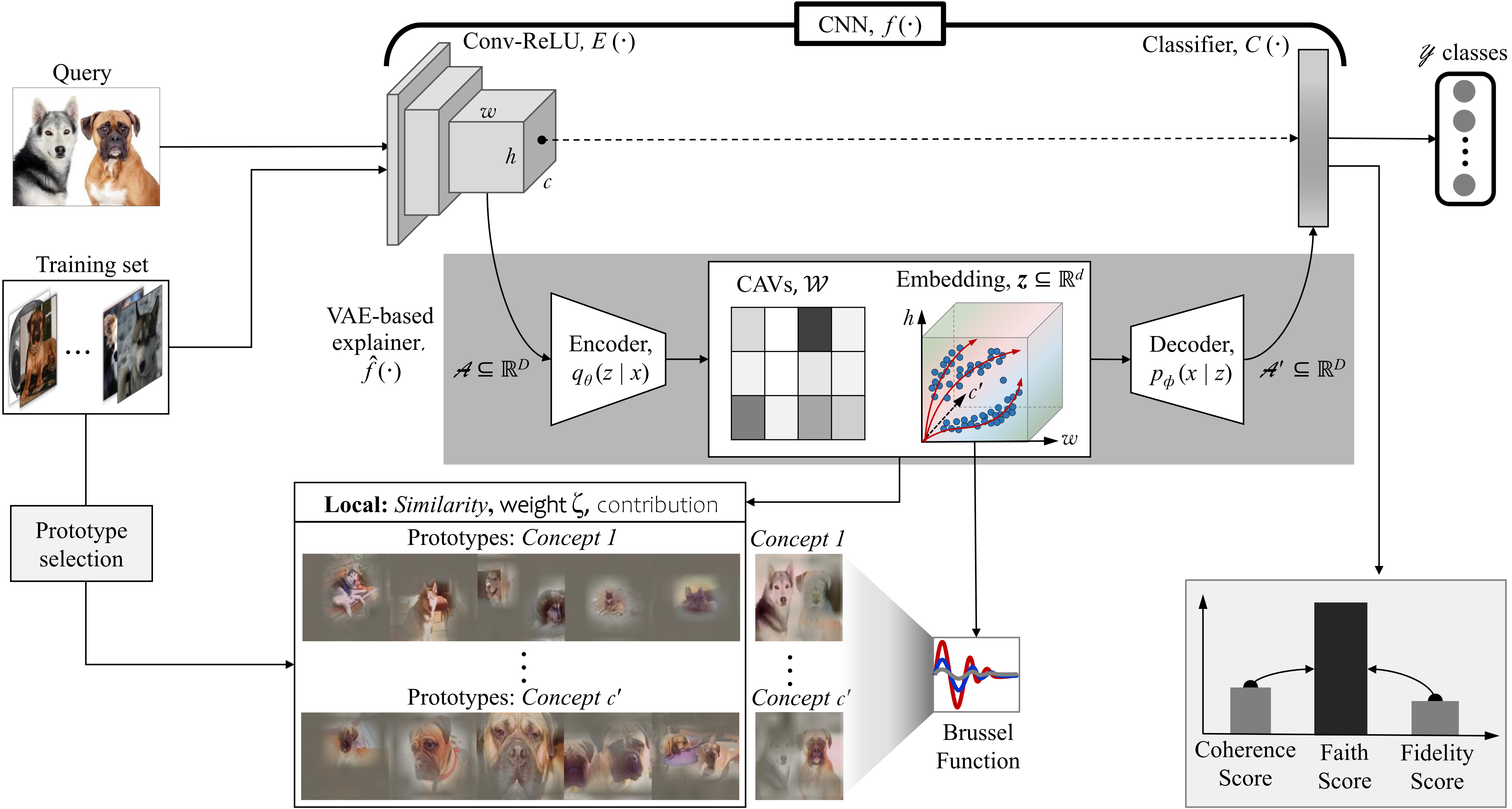}
\caption{The proposed TraNCE explainability framework. The CNN $f(\cdot)$ comprises Conv-ReLU layers $E(\cdot)$, generating high-dimensional activations $\mathcal{A}_{l}$ at layer $l$, and a classifier $C(\cdot)$ for predictions. The VAE-based explainer $\hat{f}(\cdot)$ includes an encoder and decoder, producing CAVs $\mathcal{W}$ and embedding $z$ from $\mathcal{A}_{l}$ at the bottleneck. Concepts are derived from embedding and visualized using the Bessel function, with prototypes sourced from the training set via MMD-Critic. Each concept explanation uses $\mathcal{W}$ to compute local checkpoints: concept \textit{Similarity}, concept weights $\zeta$, and concept \texttt{Contribution}. Global explanations are provided by $C(\cdot)$, evaluating prediction losses between $\mathcal{A}_{l}$ and $\mathcal{A}^\prime_{l}$ as \textit{Faith} scores.}
\label{sys_model}
\end{figure*}

Fig.~\ref{sys_model} illustrates TraNCE schematically, with detailed pseudo-code in Algorithm~\ref{alg:algorithm}. The CNN $f(\cdot)$ consisting of $E(\cdot)$ that generates $\mathcal{A}_{l}$ and $C(\cdot)$ for class predictions can be split at the intermediate layer such that $f(x) \equiv E(x) | C(\mathcal{A}_{l})$ (step 1 of Algorithm~\ref{alg:algorithm}). Unsupervised concept discovery from the high-dimensional $\mathcal{A}_{l}$ involves employing an explainer for decomposing $\mathcal{A}_{l}$ to lower-dimensional embedding $z \in \mathbb{R}^d$ for explanation. Since the information in $\mathcal{A}_{l}$ is complex with non-linear relationships, the explainer's objective is to extract meaningful information, simplify its complexity with minimal information loss, offer qualitative explanations, and provide quantitative explanations via reliable quantitative checkpoints. 

As shown in Fig. \ref{sys_model}, the VAE-based explainer $\hat{f}(\cdot)$ includes an encoder and decoder, producing the latent embedding $z$ and CAVs $\mathcal{W}$ from $\mathcal{A}_{l}$ at its bottleneck after self-supervised training such that $\hat{f}(\mathcal{A}_L) \approx z \cdot \mathcal{W}$ (steps 2--9 of Algorithm~\ref{alg:algorithm}). User-defined $c^\prime$ number of concepts are derived from embedding and visualized using the Bessel function, which interpolates pixel activations to create smooth transitions between high and low activations. Prototypes for each concept are sourced from the training set via the MMD-Critic, which automatically retrieves images with high activations, highlighting discovered concepts with the Bessel function. Each concept (and prototype) explanation exploits $\mathcal{W}$ for the local quantitative explanation which includes the \textit{Similarity} of the concepts to the target class and their \texttt{Contribution}s to the concept explanation, quantified as the concept importance (step 10--12 of Algorithm~\ref{alg:algorithm}). Using $C(\cdot)$, the global faithfulness of concept explanations can be evaluated by comparing the prediction losses between $\mathcal{A}_{l}$ and the explainer's inverted activations $\mathcal{A}^\prime_{l}$, resulting in the Fidelity, Coherence and comprehensively, \textit{Faith} scores (step 13--17 of Algorithm~\ref{alg:algorithm}).

\begin{algorithm}[tb]
\caption{Pseudo-code for TraNCE}
\label{alg:algorithm}
\textbf{Input}: Image ($x$), CNN model $f(x)$, VAE model $\hat{f}(x)$\\
\textbf{Parameter}: User-defined $c^\prime$, epochs $\rho_{max}$\\
\textbf{Output}: \texttt {Contribution} and \textit{Faith} score. \\ 
\vspace{-.3cm}

\begin{algorithmic}[1] 
\STATE Split $f(x) \equiv E(x) | C(\mathcal{A}_{l})$ such that $\mathcal{A}_{l} \in \mathbb{R}^{D}$ in layer $l$;

\STATE Reshape $\mathcal{A}_{l}$ to $\mathcal{G}_{l} \in \mathbb{R} ^ {(h \cdot w) \times c}$;

\STATE Normalize $\mathcal{G}_{l} \sim x(0,1) \in \mathbb{R}^{(h \cdot w) \times c}$;\

\FORALL{$a^{(i,j)}$ in $x$}

    \WHILE{$\rho_{max}$}
    
        \STATE Train $\hat{f}_{\{\theta, \phi\}}(x, \hat{x}): x \stackrel{q_\theta(z \mid x)}{\longrightarrow} z \stackrel{p_\phi(x \mid z)}{\longrightarrow} \hat{x}$ such that
        
        \STATE $\hat{f}_{\{\theta, \phi\}}(x, \hat{x})  \equiv \{\hat{f}_{\theta}(x, z), \hat{f}_{\phi}(z, \hat{x})\} = \mathcal{W}_l(x) +b$;
        
        \STATE Save $\hat{f}_{\{\theta, \phi\}}(x, \hat{x}) \text{ and } \mathcal{W}_l$;
    \ENDWHILE
    
    \STATE Create heatmap $\mathcal{Z}_{J_{v}}$ from $z$ using Eq.~\ref{enhance_heatmap};
    
    \STATE Overlay $\{z_i \in \mathcal{Z}_{J_{v}} \mid x \in \mathcal{X}\}$;
    
    \STATE Compute \texttt{Contribution} = \textit{Similarity} $\cdot  \mathcal{\zeta}$; 
    
\ENDFOR

\STATE  Invert $\hat{x}(0,1) \in \mathbb{R}^{(h \cdot w) \times c} \longleftarrow z \in \mathbb{R}^{(h \cdot w) \times c^\prime}$;

\STATE Denormalize $\hat{x}(0,1) \sim \mathcal{G} ^{\prime}_l \in \mathbb{R}^{(h \cdot w) \times c}$;\

\STATE Reshape $\mathcal{G} ^{\prime}_l$ to $\mathcal{A} ^{\prime}_l \in \mathbb{R}^{(h \cdot w) \times c}$;

\STATE Compute $\textit{Faith}$;

\STATE \textbf{return} $\texttt {Contribution}$ and $\textit{Faith}$.
\end{algorithmic}
\end{algorithm}

\subsection{VAE-based Decomposition}

A reducer's decomposition efficiency is crucial for the explainer's success, requiring minimal information loss \cite{ice, craft, pca_variants, vae_explain, onceclassVAE}. Linear reducers fail to capture the intrinsic geometric structure of $\mathcal{A}_{l}$. In contrast, the VAE's probabilistic nature avoids the pitfalls of linear reconstruction assumptions, its adaptive nature allows for case-specific adjustments, and its continuous latent space facilitates smooth transitions between data points \cite{vae_explain}.

The VAE employs a probabilistic self-supervised encoder-decoder neural network with trainable parameters $\theta$ (encoder) and $\phi$ (decoder) to generate a lower-dimensional latent embedding from a high-dimensional input~\cite{vae_explain, onceclassVAE}. To achieve this, the multidimensional $\mathcal{A}_{l}$ shape ($h, w, c$) is first reshaped across $c$: $\mathcal{A}_{l} \rightarrow \mathcal{G}_{l} \in \mathbb{R}^{(h \cdot w) \times c}$ to form a matrix,  each row in $\mathcal{G}_{l}$ corresponding to a specific spatial location in the $\mathcal{A}_{l}$. Next, the encoder, defined by the standard Gaussian prior, encodes the normalized $\mathcal{G}_{l} \sim x(0,1) \in \mathbb{R}^{(h \cdot w) \times c}$ to generate $z$ at the bottleneck such that the variational posterior for an observation is a corresponding normal distribution $q_\theta(z \mid x)= \mathcal{N}\left(z_j \mid \mu_j(x), \sigma_j^2(x)\right)$ with the mean and variance produced by the encoder. The decoder (also probabilistic) decodes $z$ to reproduce the input $\hat{x}$ such that $p_\phi(\hat{x} \mid z)= \mathcal{N}\left( \mu_j(\hat{x}), \sigma_j^2(\hat{x}) \mid z_j\right)$  \cite{vae_explain}. 
 
\begin{definition}
    \textit{The non-negative latent representation $z \in \mathbb{R}^d$ of the high-dimensional feature map $x \in \mathbb{R}^D$ is generated by the reducer (encoder) $\hat{f}_{\theta}(x, z)$ of a trained VAE $\hat{f}_{\{\theta, \phi\}}(x, \hat{x})$ as the rectified decomposition of $x$ from the $D$ to $d$ dimension.}
\end{definition}

Practically, $\hat{f}_{\{\theta, \phi\}}(x, \hat{x})$ aims to reconstruct $x \approx \hat{x}$ with a minimized, regularized reconstruction loss. Evaluating the efficiency of the decoder in reconstructing $x$ from $z$ can be reliably assessed by minimizing the decomposition loss in a regularized manner ensuring that meaningful embedding are produced by diversifying each $z_i$ without isolating data points in separate regions of Euclidean space \cite{vae_explain, onceclassVAE}.

Training $\hat{f}_{\{\theta, \phi\}}(x, \hat{x})$ (steps 3 to 8 of Algorithm \ref{alg:algorithm}) involves iterating stochastic gradient descent to optimize the regularized reconstruction loss with respect to $\theta$ and $\phi$. Normalizing before training enhances inference and model stability \cite{vae_explain, automate}. For a step size $\rho_i (i = 1,2,3, \ldots, max)$, the parameters are updated thus: $\theta \leftarrow \theta-\rho_i \frac{\partial l}{\partial \theta}$,  $\phi \leftarrow \phi-\rho_i \frac{\partial l}{\partial \phi}$ to derive the trained explainer $\hat{f}_{\{\theta, \phi\}}(x, \hat{x}): x \stackrel{q_\theta(z \mid x)}{\longrightarrow} z \stackrel{p_\phi(x \mid z)}{\longrightarrow} \hat{x}$ such that:

\subsection{Automatic Concept Discovery and Visualization}

Concepts are derived from $z$ and visualized using the Bessel function, which interpolates pixel activations to create smooth transitions between high and low activations. The latent embedding $z$ can then be obtained by a non-linear transformation by the trained encoder: 
\begin{equation}
    z= \hat{f}_{\theta}(x, z)\mathcal{A}_{l} = ReLU(\mathcal{W}_l (x) + b),
\end{equation}

\begin{prop}
    \textit{A trained VAE $\hat{f}_{l_{i}(\theta, \phi)_{min}}(x, \hat{x})$ consists of a reducer $\hat{f}_{\theta_{min}}(x, z)$ and a reconstruction model $\hat{f}_{\phi_{min}}(z, \hat{x})$. The fixed weight matrix $\mathcal{W}_l \in \mathbb{R}^{c^\prime \times c}$ and the embedding $z \in \mathbb{R}^{(h \cdot w) \times c^\prime}$ produced at the bottleneck represent CAVs and concepts in the feature map space respectively. }
\end{prop}

For new images, the reducer $\hat{f}_{\theta_{min}}(x, z) \approx z \cdot \mathcal{W}$ generates latent embedding $z$ with the CAVs $\mathcal{W}$ as the degree to which vector $a^{(i,j)}$ is similar (\texttt{Similarity score}) to the CAVs in $\mathcal{W}_l$. Deriving the concepts from $z$ entails mapping each basis pattern in $z$ back to its original spatial dimension i.e. reshape $z \in \mathbb{R}^{(h \cdot w) \times c^{\prime}} \rightarrow \mathcal{Z} \in \mathbb{R}^{h \times w \times c^{\prime}}$ to produce $c^\prime$ heatmaps corresponding to distinct spatial concepts with dimensions ($h, w$). Each reshaped spatial component $\mathcal{Z}[:, :, i] \in \mathbb{R}^{h \times w}$ represents a distinct concept that the CNN has learned. This forms a matrix which can be enhanced using Eq.~\ref{besss} to form:
\begin{equation}
    \mathcal{Z}_{J_{v}} = \mathcal{Z}[:, :, i] \cdot J_v(\mathcal{Z})
    \label{enhance_heatmap}
\end{equation}
The enhanced heatmap is then resized to the original image size and overlayed on the image as $\{z_i \in \mathcal{Z}_{J_{v}} \mid x \in \mathcal{X}\}$, revealing the concepts as high activations (step 10--11 of Algorithm~\ref{alg:algorithm}). 

\subsection{Concept Contribution Estimation}
Computing the directional derivative of the linear classifier $C(\cdot)$ to the learned CAV $\mathcal{W}_l$ in $\mathcal{A}_l$ helps estimate the sensitivity of the extracted CAVs \cite{tcav}. $C(\mathcal{A}_l)$ is linear with trainable weights $\textbf{t} \in \mathbb{R}^c$ that ensure consistency of the derivatives across the entire input space. 

\begin{definition}
    \textit{Given the image activation $\mathcal{A}_l$, a logit function $h_{l, y_i}: \mathbb{R}^{y_i} \rightarrow \mathbb{R}$, and a learned CAV $\mathcal{W}_l$ at layer $l$, the conceptual sensitivity for a target class $y_i$ is given by} 
$\frac{\partial C_{l, y_i}}{\mathcal{W}_l}=  \lim _{\epsilon \rightarrow 0} \frac{h_{l, y_i}\left(\mathcal{A}^D_l+\epsilon \mathcal{W}_l\right)-h_{l,y_i}\left(\mathcal{A}^D_l\right)}{\epsilon}$, \textit{at minimal additive regularization $\epsilon$} \cite{tcav}.

\end{definition}

The estimated concept \texttt{weight} is calculated by averaging the derivatives $\frac{\partial C_{l, y_i}}{\mathcal{W}_l}$ over $\mathcal{A}_l$. Consequently, the concept importance of CAV $\mathcal{W}_l$ is the product of the classifier weight $\textbf{t}$ and the $\mathcal{\zeta}$ i.e \texttt{Contribution}  = $\textbf{t} \cdot \mathcal{\zeta}$ , following a global average pooling of $\mathcal{A}_l$. Each concept's \texttt{Contribution} indicates how much the specific concept influences the CNN's decision, invariably answering the question: \textit{how much does the presence or absence of a specific concept influence the CNN's decision?}

\subsection{Prototype Selection}

Prototypical explanations provide concrete examples that illustrate how the CNN interprets and classifies different concepts, enhancing the transparency and interpretability of the model's decision-making process. 
A feature map may be considered a vector representation of a conceptual part (e.g. eye, mouth, ear). Thus, factorizing these vectors can disentangle relevant and frequently occurring CAVs. These frequently occurring CAVs guide the selection of concept prototypes i.e. regions on images containing the target concepts from the training set whereby the images with frequently occurring CAVs are selected as the prototypes.

Our objective is to generate prototypes $\mathcal{X}_{s}=\left\{x_i \forall i \in \mathcal{S}; \mathcal{S} \in[n]\right\}$ from training images that faithfully represent a concept in images $\mathcal{X}=\left\{x_i, i \in[n]\right\}$. MMD-critic achieves this by selecting prototype indices $\mathcal{S}$ which minimize their square of the maximum mean discrepancy  $\operatorname{MMD}^2\left(\mathcal{F}, \mathcal{X}, \mathcal{X}_{\mathcal{S}}\right)$ by minimizing:
\begin{equation}
   J_b(\mathcal{S}) =\frac{1}{n^2} \sum_{i, j=1}^n k\left(x_i, x_j\right)-\mathrm{MMD}^2\left(\mathcal{F}, \mathcal{X}, \mathcal{X}_{\mathcal{S}}\right),
    \label{mmd}
\end{equation}
where $\mathcal{F}$ is a reproducing kernel Hilbert space (RKHS) with kernel function $k(\cdot, \cdot)$. The MMD-critic then selects $m_*$ prototypes as the subset of indices $\mathcal{S} \in[n]$ which optimize $\max _{\mathcal{S} \in 2^{[n],|\mathcal{S}| \leq m_*}} J_b(\mathcal{~S})
 \label{mmd2}$.

\subsection{Faithfulness Evaluation}

To address the limitations inherent in the Fidelity-only paradigm for global faithfulness evaluation \cite{MAPE}, we introduce Coherence, which assesses concept consistency by computing the normalized cross-spectral density of the explainer's predictions \cite{randomdata}. 

\begin{prop}
   \textit{ An explainer is coherent if the set of prototypes it discovers for an image class is consistent with the distinguished regions of the image.}
\end{prop}

Given the predictions from the CNN and explainer: $f(I)$ and $\hat{f}(I)$ where $\hat{f}_l(i) = C_l\left(f_{\phi}(z, \hat{x})\left(f_{\theta}(x, z)\left(E_l(I)\right)\right)\right)$, compute their finite Fourier transform respectively at a sampling rate of $\mathbb{S}$ to obtain $F(s)$ and $\hat{F}(s)$. For any value of $s$ ($0 \leq s \leq \mathbb{S}$), the one-sided cross-spectral density function $G_{f\hat{f}}(s)$ is expressed in terms of the magnitude $|G_{f\hat{f}}(s)|$ and phase $e^{-j \theta_{f\hat{f}}(s)}$ as: 
\begin{equation}
G_{\hat{f}f}(s) = G_{f\hat{f}}^*(s)=\left|G_{f\hat{f}}(s)\right| e^{-j \theta_{f\hat{f}}(s)},
\label{cross_phase}
\end{equation}
For any real constants $a$ and $b$, the squared absolute sum of $F$ and $\hat{F}$ will be greater than or equal to zero such that $\left|a F(s)+b \hat{F}(s) e^{j \theta_{f\hat{f}}(s)}\right|^2 \geq 0$. Expanding and taking the expectation of Eq. (\ref{cross_phase}) over an arbitrary index, multiplying by ($2\mathbb{S}$), and
letting $\mathbb{S}$ decrease without bound, 
\begin{equation}
\begin{array}{l}
a^2 G_{ff}(s)+\\
a b\left[G_{f\hat{f}}^*(s) e^{j \theta_{f\hat{f}}(s)}+G_{\hat{f}f}(s) e^{-j \theta_{f\hat{f}}(s)}\right]\\
+b^2 G_{\hat{f}\hat{f}}(s) \geq 0,
\end{array}
\label{eq_expand_cross}
\end{equation}

Comparing the middle term of Eq. (\ref{eq_expand_cross}) and Eq. (\ref{cross_phase}) shows that $G_{f\hat{f}}^*(s) e^{j \theta_{f\hat{f}}(s)}+G_{\hat{f}f}(s) e^{-j \theta_{f\hat{f}}(s)}=2\left|G_{f\hat{f}}(s)\right|$; thus, Eq. (\ref{eq_expand_cross}) is reduced to $a^2 G_{ff}(s)+2 a b\left|G_{f\hat{f}}(s)\right|+b^2 G_{\hat{f}\hat{f}}(s) \geq 0$, a quadratic equation with non-positive roots $a/b$ and a non-positive discriminant: $4 \left|G_{f\hat{f}}(s)\right|^2 - 4 G_{ff}(s)G_{\hat{f}\hat{f}}(s) \leq 0$, such that $\left|G_{f\hat{f}}(s)\right|^2 \leq G_{ff}(s)G_{\hat{f}\hat{f}}(s) $. Consequently, the coherence function (sometimes called the coherence squared function) may be defined as the ratio: 
\begin{equation}
\begin{array}{ll}
\gamma_{f, \hat{f}}^2(s)=\frac{\left|G_{f\hat{f}}(s)\right|^2}{G_{ff}(s) G_{\hat{f}\hat{f}}(s)}, & 0 \leq \gamma_{f, \hat{f}}^2(s) \leq 1,
\label{cohe_}
\end{array}
\end{equation}
where $G_{f(s)}$ and $G_{\hat{f}(s)}$ are the autospectral densities of the CNN's and reducer's predictions respectively while $G_{f(s)\hat{f}(s)}$ is the cross-spectral density between the predictions. The supplementary material contains the derivation of Eq. \ref{cohe_}. Fidelity as proposed by \cite{why_trust_you} is defined as:
\begin{equation}
\begin{array}{ll}
Fid_{f, \hat{f}}(I)=\frac{\sum_{i \in I}|f(i)-\hat{f}(i)|}{\sum_{i \in I}|f(i)|} & 0 \leq Fid_{f, \hat{f}}(I) \leq 1.
\label{Fidelity_eqn}
\end{array}
\end{equation}

\begin{prop}
\textit{The average sum of Coherence $Coh_{F, \hat{F}}$ and Fidelity $Fid_{F, \hat{F}}$, $\textit{Faith} = (Fid_{f, \hat{f}} +\gamma_{f, \hat{f}}^2)/2$ offers a comprehensive faithfulness evaluation paradigm, leveraging the strengths of one to address the weaknesses of the other.}
\end{prop}

Combining these insights allows for considering trade-offs between the metrics, ensuring a well-rounded evaluation resilient to biases or fluctuations in individual measures~\cite{cam_metrics}. This approach avoids overemphasizing a single metric and contributes to a nuanced understanding of the explainer's faithfulness. The proposed \textit{Faith} metric facilitates clearer insights for informed decision-making. It encourages the development of explainers that can be quantitatively assessed for faithfulness and sanity, reducing the chances of human bias.

\section{Experiments}
\label{experiments}

We conducted experiments on Torchvision's pre-trained ResNet \cite{resnet,wideresnet}, ResNext \cite{resnext} and InceptionV3 \cite{inceptionv3} models using the ILSVRC2012 \cite{imagenet} dataset. Table \ref{table2} shows the top-1 test accuracies of the pre-trained CNN models. The results demonstrate good classification performance, with ResNet50 and InceptionV3 outperforming the other models.

\begin{table}[!t]
\begin{center}
\caption{Top-1 accuracy of CNN models on the ILSVRC2012 dataset.}
\label{table2}
\begin{tabular}{cc}
\hline
CNN Model & Top-1 accuracy\\
\hline
ResNet18 \cite{resnet}  & 76.8\% \\
ResNet50 \cite{resnet} & \underline{\textit{78.9\%}}\\
ResNext101 \cite{resnext}  & 77.6\% \\
WideResNet50 \cite{wideresnet} &  77.8\% \\  
WideResNet101 \cite{wideresnet}  &  78.5\% \\
InceptionV3 \cite{inceptionv3} & \textbf{80.2\%} \\
\hline
\end{tabular}
\end{center}
\end{table}


\begin{figure}[H]
\begin{minipage}[b]{0.49\linewidth}
  \centerline{\includegraphics[width=1\linewidth]{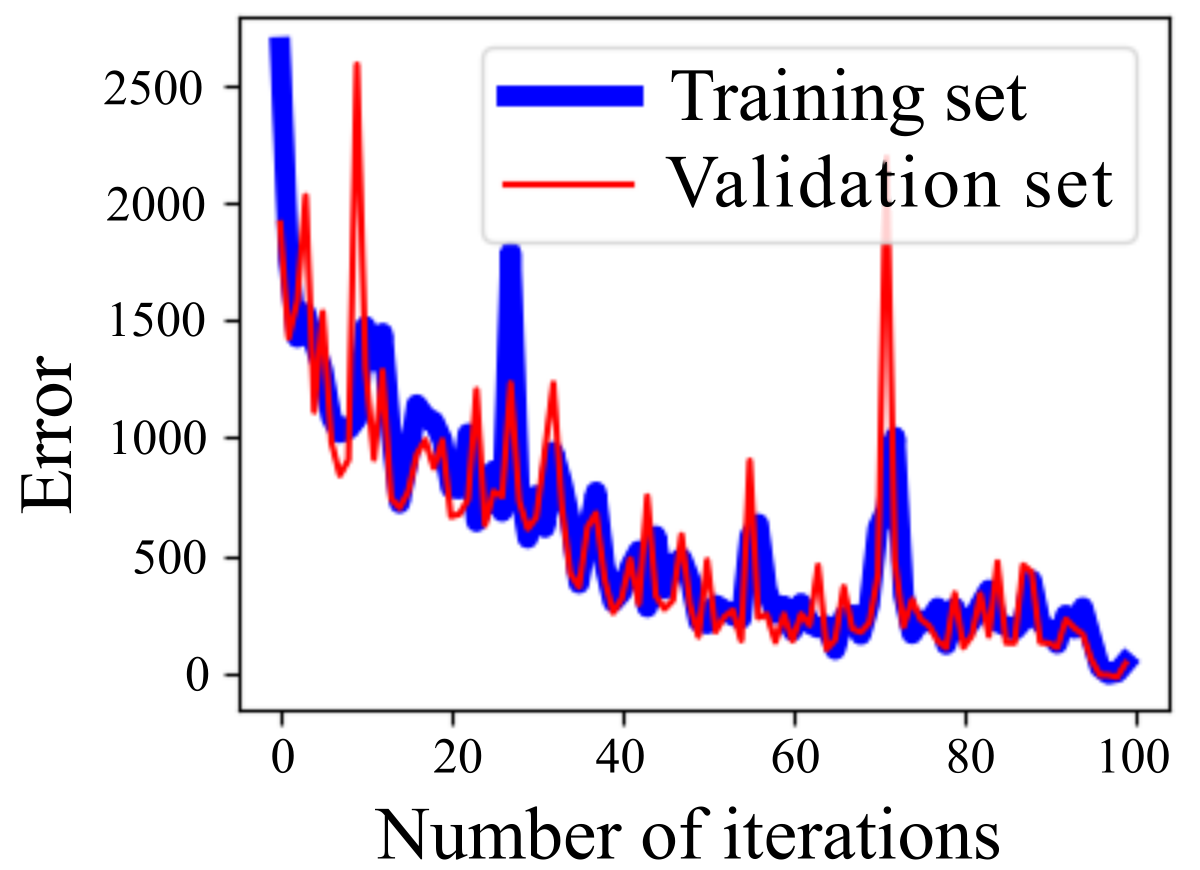}}
  \centerline{(a)}\medskip
\end{minipage}
\begin{minipage}[b]{.49\linewidth}
  \centerline{\includegraphics[width=1\linewidth]{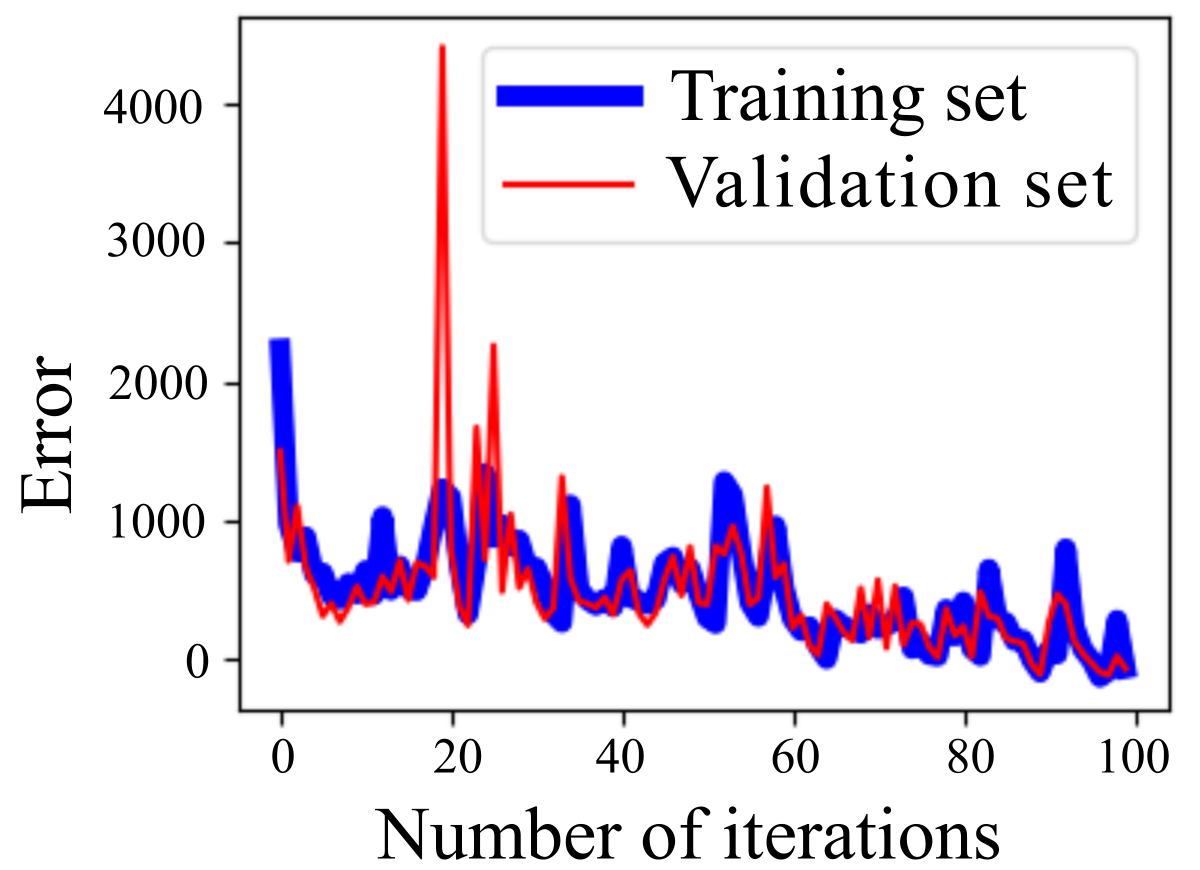}}
  \centerline{(b)}\medskip
\end{minipage}
\caption{Train/validation history of the proposed TraNCE explainer over 100 iterations for (a) ResNet50's explanation using an image of an Australian Kelpie and a Chihuahua, and (b) InceptionV3's explanation using an image of a Macaw and an Eagle.}
\label{training}
\end{figure}

\subsection{Training the TraNCE Explainer}

Designed to quantify the difference between the true and transformed image activations i.e. $\mathcal{A}_l$ and $\mathcal{A}^{\prime}_l$ respectively, the training objective for the VAE is to train $\mathcal{A}_l$ iteratively in a self-supervised manner. First, we target each model's intermediate layer: $layer4$ for ResNet18, ResNet50 \cite{resnet}, ResNext101 \cite{resnext}, WideResNet101, and WideResNet50 \cite{wideresnet} respectively and the $Mixed\_7c$ layer for the InceptionV3 \cite{inceptionv3}. Next, we designed a seven-layer VAE model with the encoder dimensions {$c$, $c/2$, $c/4$, $c^\prime$} and symmetric decoder dimensions {$c^\prime$, $c/4$, $c/2$, $c$}. Such simple architectures reduce computation times and overfitting \cite{vae_explain, onceclassVAE}. 

We deployed each CNN on class-specific images, producing unique $\mathcal{A}_l$ at intermediate layers. For each $\mathcal{A}l$, we trained a VAE using stochastic gradient descent, optimized with Adam (initial learning rate 0.005) and a Cosine annealing scheduler over 100 iterations, with early stopping at $l_i(\theta, \phi){min}$ in batches of 16. Cross-validation ensured a well-trained $\hat{f}{\theta}(x, z)$ and evaluated model reliability, mitigating accidental success and overfitting/underfitting. Each VAE model was saved for inference, including the encoder, decoder, and weights. Fig. \ref{training} shows training histories (training and validation losses) over 100 iterations for two local explanation examples: ResNet50 with images of an Australian Kelpie and a Chihuahua and InceptionV3 with images of a Macaw and an Eagle (see Fig. \ref{local_explanations} for their full concept explanations). In these examples, $c^\prime$ = 32 as suggested in \cite{overlookedfactors}. 

As shown in Fig. \ref{training}, the training history reveals good loss convergence for both the training and validation sets, suggesting minimal differences between $\mathcal{A}_l$ and $\mathcal{A}^{\prime}_l$. This implies that $z$ at the bottleneck provides a representative latent embedding of $\mathcal{A}_l$.

\subsection{Automatic Concept Discovery and Visualization}

For a class of interest, such as an Australian Kelpie, TraNCE generates image embedding from the training dataset and, using the concept visualization module, displays concepts by highlighting image regions that represent a concept in the target class and masking the rest. Most existing visualization methods use a threshold-based approach, highlighting pixels with activations above a certain threshold and masking the rest. While effective, this approach presents challenges: selecting an optimal threshold introduces human bias \cite{ice}, the choice of data augmentation techniques affects the quality of the concept bank \cite{craft}, and issues of concept duplication and ambiguity remain \cite{ace, ice, ncaf}. (see Section \ref{sanity}). To address these challenges, we use the Bessel function \cite{bessel} with the MMD-critic for prototype selection and criticism \cite{example_not_enough}, interpolating pixel activations to create smooth transitions between high and low concept activations. We chose a red-blue-dark grey color map with 40\% transparency for enhanced qualitative explanations: red highlights strong concept representations, grey indicates minimal representations, and blue shows moderate representations. This smooth transition in the heatmap provides a more intuitive visualization of discovered concepts and prototypes. Unlike threshold-based methods, which can lead to ambiguous and duplicate concepts, our approach accurately reveals what the CNN saw and activation strengths across pixel values for the target class. Bessel interpolation ensures the heatmap reflects the underlying data distribution while minimizing artefacts like jagged edges or abrupt transitions. 

When provided with a query image, the trained explainer generates the image embedding as concepts accompanied by prototypes. These provide qualitative explanations, while quantitative explanations are given as \texttt{Weight}, \texttt{Contributions}, and \textit{Faith} scores. Fig. \ref{local_explanations} shows local explanations for ResNet50 using images of an Australian Kelpie and a Chihuahua, and InceptionV3 using images of a Macaw and an Eagle. As shown in Fig. \ref{local_explanations}, the discovered concepts are accompanied by five prototypical images alongside their similarity scores with the concept and their \texttt{Weight} and \texttt{Contribution} scores respectively. Comprehensively, the total \texttt{Contribution} scores for each target class are compiled as the sum of each concept's \texttt{Contribution}, aligning closely with the CNN's prediction and providing a human-understandable probability of what the query image looks like. 

While it's impractical to present explanations for all 1000 classes in the ILSVRC2012 dataset, the discovered concepts for tested classes align well with associated prototypes for the tested examples. We use the Bessel function to represent pixel activations, with grey indicating low contribution, blue for moderate, and red for high. This method provides a realistic visualization of identified concepts and prototypes, revealing not only where the CNN looked but also what it saw and the significance of these perceptions as \texttt{Weight} and \texttt{Contribution} scores. Combining qualitative and quantitative explanations offers a more comprehensive understanding than relying solely on visual activations.

\begin{figure}[!t]
    \begin{minipage}[b]{.49\linewidth}
      \centerline{\includegraphics[width=1\linewidth]{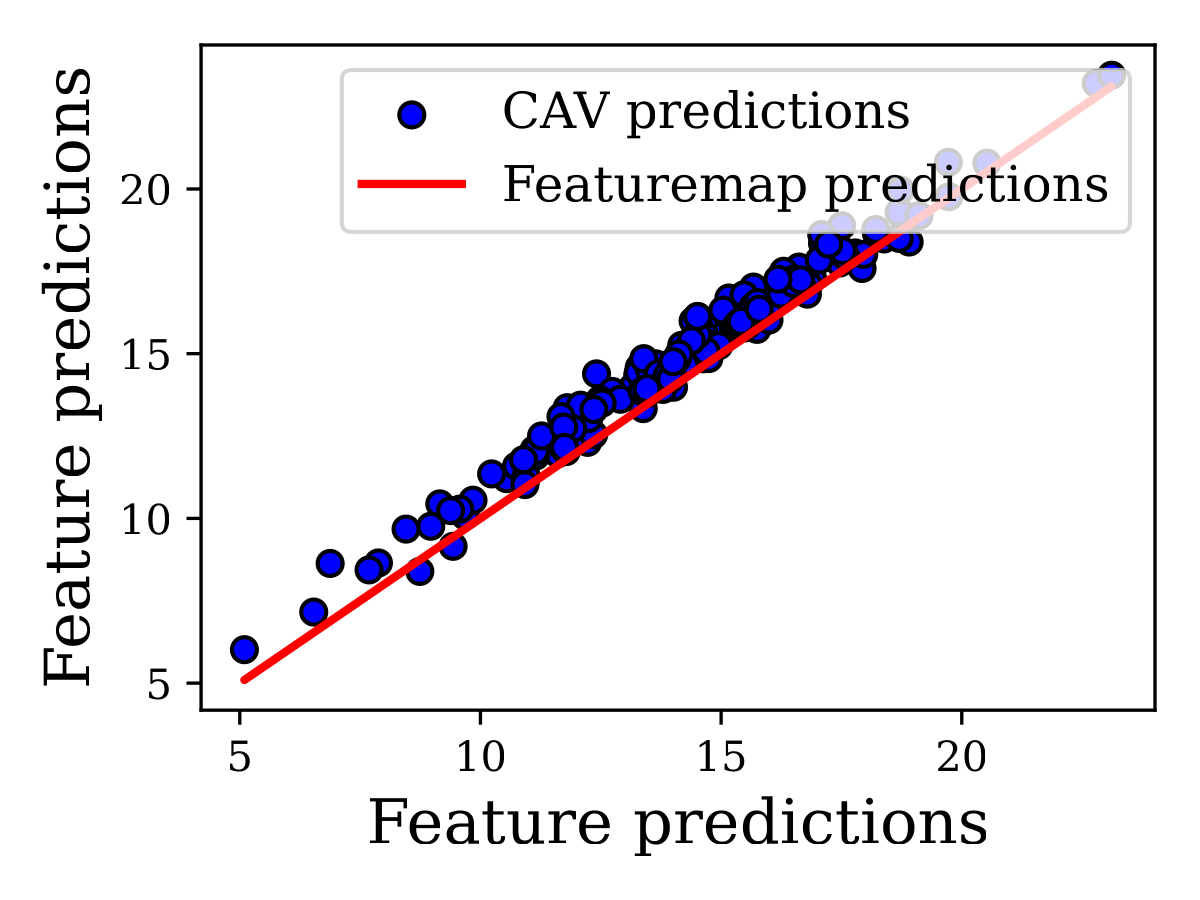}}
      \vspace*{-12ex}  \hspace*{2ex}  
        \begin{center}
       {\hspace*{7ex} \small \textit{Faith} = $98.9\pm1.0\%$  }
        \end{center}\vspace*{3mm}\medskip\vspace*{2mm}
        \centerline{(a)} \medskip
        \vspace*{12ex} \hspace*{.2ex} 
    \end{minipage} 
    \begin{minipage}[b]{.49\linewidth}
      \centerline{\includegraphics[width=1\linewidth]{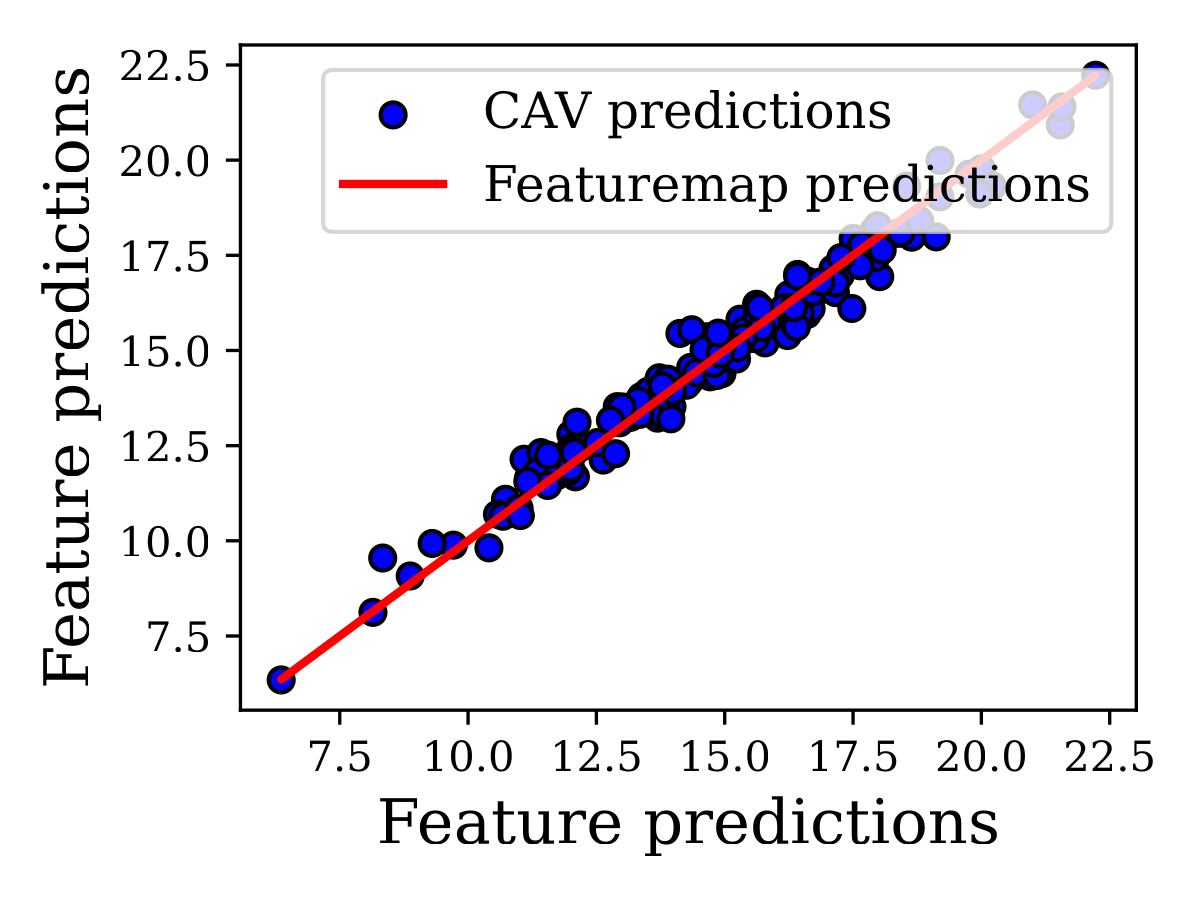}}
      \vspace*{-12ex} \hspace*{-.2ex}  
        \begin{center}
        {\hspace*{6ex} \small \textit{Faith} = $99.1\pm0.7\%$  }
        \end{center}\vspace*{3mm} \medskip \vspace*{2mm}
        \centerline{(b)}\medskip
        \vspace*{12ex} \hspace*{.2ex} 
    \end{minipage}\vspace*{-22mm}
    
    \begin{minipage}[b]{.49\linewidth}
      \centerline{\includegraphics[width=1\linewidth]{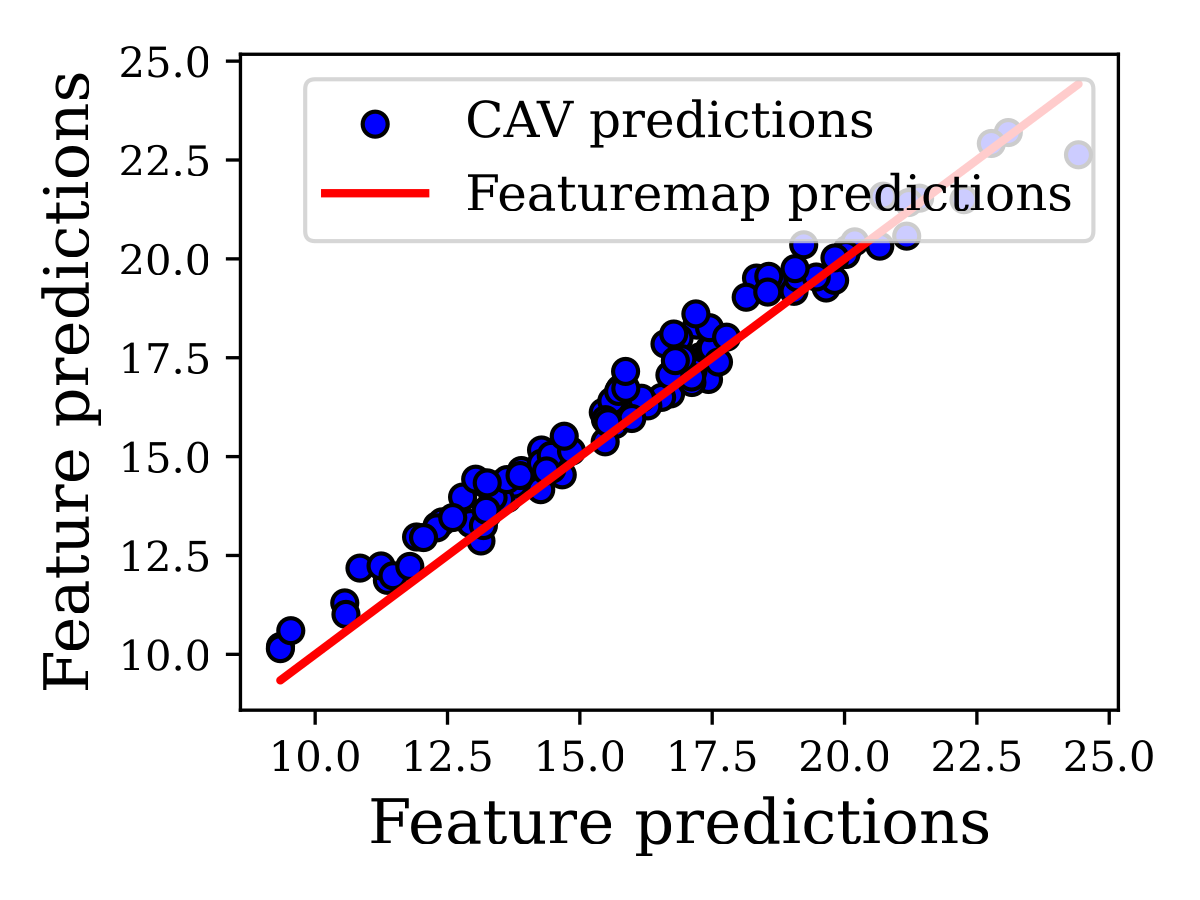}}
      \vspace*{-12ex} \hspace*{-.2ex}  
        \begin{center}
        {\hspace*{5.5ex} \small \textit{Faith} = $99.2\pm0.7\%$  }
        \end{center}\vspace*{3mm}\medskip \vspace*{2mm}
        \centerline{(c)}\medskip
        \vspace*{12ex} \hspace*{.2ex} 
    \end{minipage} 
    \begin{minipage}[b]{.49\linewidth}
      \centerline{\includegraphics[width=1\linewidth]{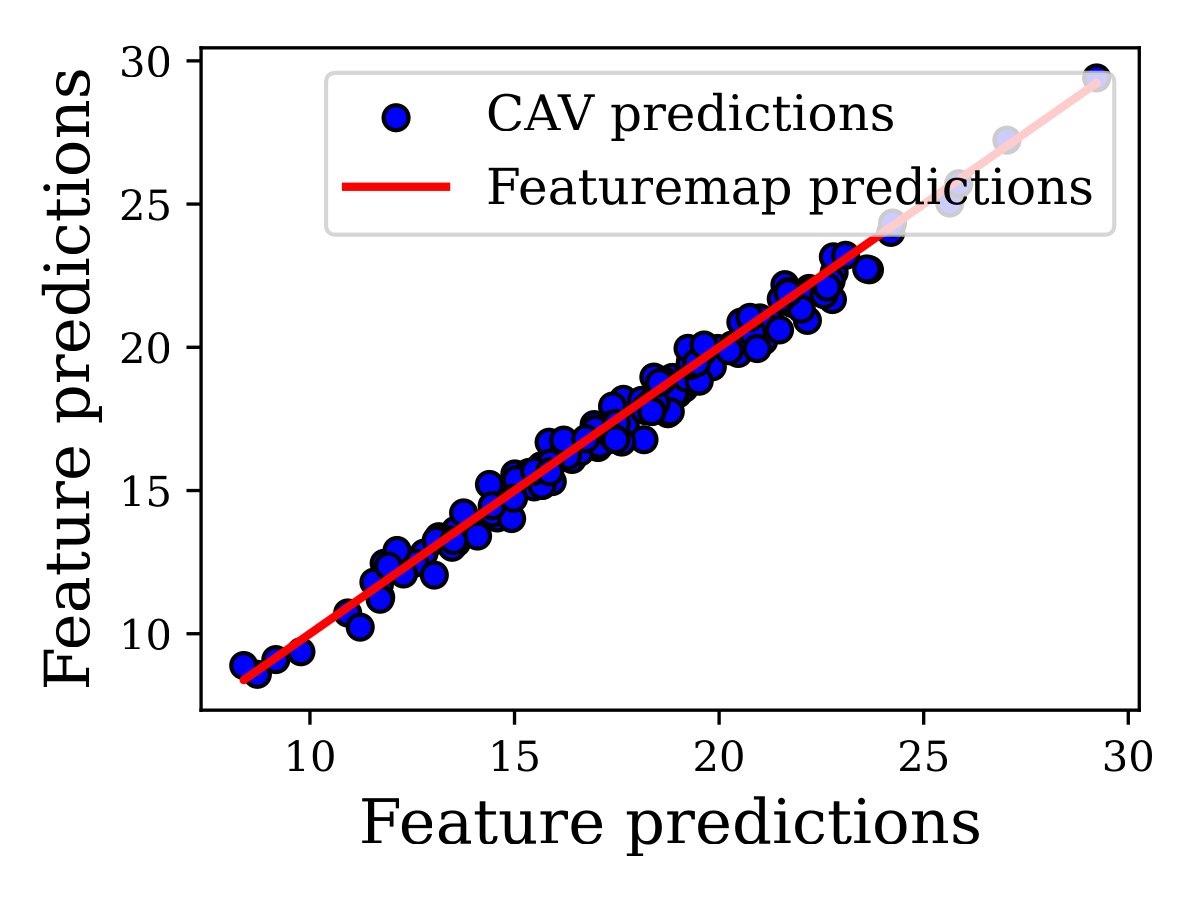}}
      \vspace*{-12ex} \hspace*{-.2ex}  
        \begin{center}
        {\hspace*{6ex} \small \textit{Faith} = $99.6\pm0.4\%$  }
        \end{center}\vspace*{3mm}\medskip \vspace*{2mm}
        \centerline{(d)}\medskip
        \vspace*{12ex} \hspace*{.2ex} 
    \end{minipage}\vspace*{-22mm}
\caption{Faithfulness visualisation results (\textit{Faith} scores) for (a) ResNet50's explanation for Australian Kelpie, (b) ResNet50's explanation for Chihuahua, (c) InceptionV3's explanation for Macaw, and (d) InceptionV3's explanation for Eagle. The respective \textit{Faith} scores are the mean of the Coherence (upper bound) and the Fidelity (lower bound).}
\label{best_faith_plot}
\end{figure}

\subsection{Explanations and Faithfulness Evaluation}

TraNCE's transformative architecture reconstructs $\mathcal{A}_l$ from $z^d$ to produce $\mathcal{A}^{\prime D}_l$ for computing the concept accuracy and consistency as \textit{Faith} scores. This allows for considering trade-offs between different metrics, ensuring a comprehensive evaluation resilient to biases or fluctuations in individual measures. For instance, concept explanations may be imperfect (low Fidelity) but accepted if consistent (high Coherence). In contrast, inconsistent concept explanations (high/low Fidelity, low Coherence) cannot be trusted. However, consistently accurate concepts (high Fidelity, high Coherence) are deemed faithful.

We investigated the \textit{Faith} scores for the 30 different scenarios and present some results in Fig. \ref{best_faith_plot}, the accompanying faithfulness visualisation results for the local explanations in Fig. \ref{local_explanations}. The horizontal and vertical axes represent the feature prediction values of the activation maps and CAVs respectively. The red diagonal lines represent the ground truth (activation map predictions) while the blue dots represent the CAV predictions against the ground truth. 

\begin{figure}[!t]
\centering
\vspace{.3mm}
\hrule
\begin{minipage}[c]{.18\linewidth}
  \centerline{\textbf{Cat}}
    \footnotesize{\centerline{Fidelity: 93\%}
    \centerline{Coherence: 99\%}
    \centerline{\textit{Faith}: 96\%}}
\end{minipage}\hspace{2mm}
\begin{minipage}[c]{.78\linewidth}
  \centerline{\includegraphics[width=1\linewidth]{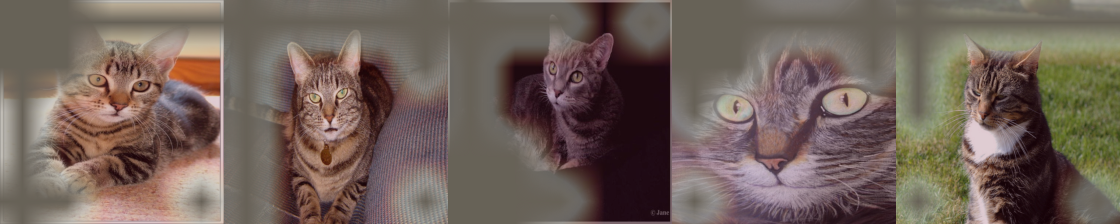}}
\end{minipage}
\hrule
\vspace{.2mm}

\begin{minipage}[c]{.18\linewidth}
  \centerline{\textbf{Birdhouse}}
    \footnotesize{\centerline{Fidelity: 96\%}
    \centerline{Coherence: 99\%}
    \centerline{\textit{Faith}: 97.5\%}}
\end{minipage}\hspace{2mm}
\begin{minipage}[c]{.78\linewidth}
  \centerline{\includegraphics[width=1\linewidth]{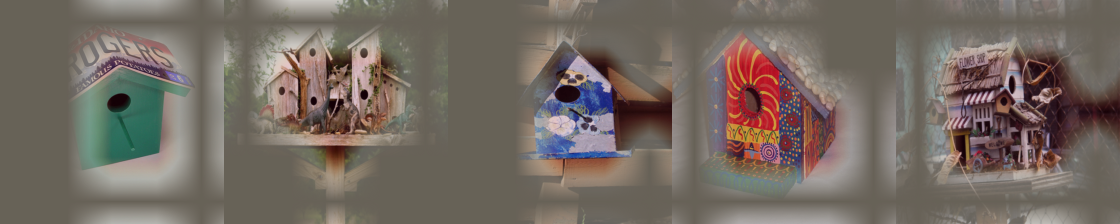}}
\end{minipage}
\hrule
\vspace{.2mm}

\begin{minipage}[c]{.18\linewidth}
  \centerline{\textbf{Buting}}
    \footnotesize{\centerline{Fidelity: 97\%}
    \centerline{Coherence: 100\%}
    \centerline{\textit{Faith}: 98.5\%}}
\end{minipage}\hspace{2mm}
\begin{minipage}[c]{.78\linewidth}
  \centerline{\includegraphics[width=1\linewidth]{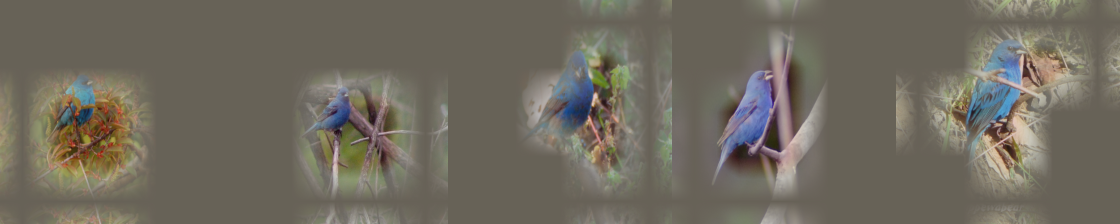}}
\end{minipage}
\hrule
\vspace{.2mm}

\begin{minipage}[c]{.18\linewidth}
  \centerline{\textbf{Cassette}}
    \footnotesize{\centerline{Fidelity: 97\%}
    \centerline{Coherence: 98\%}
    \centerline{\textit{Faith}: 97.5\%}}
\end{minipage}\hspace{2mm}
\begin{minipage}[c]{.78\linewidth}
  \centerline{\includegraphics[width=1\linewidth]{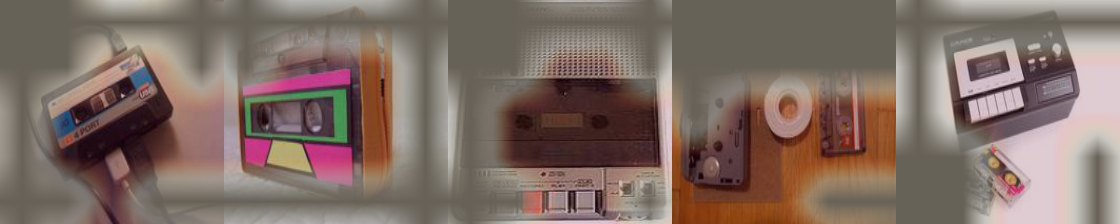}}
\end{minipage}
\hrule
\vspace{.2mm}

\begin{minipage}[c]{.18\linewidth}
  \centerline{\textbf{Sunglasses}}
    \footnotesize{\centerline{Fidelity: 96\%}
    \centerline{Coherence: 99\%}
    \centerline{\textit{Faith}: 97.5\%}}
\end{minipage}\hspace{2mm}
\begin{minipage}[c]{.78\linewidth}
  \centerline{\includegraphics[width=1\linewidth]{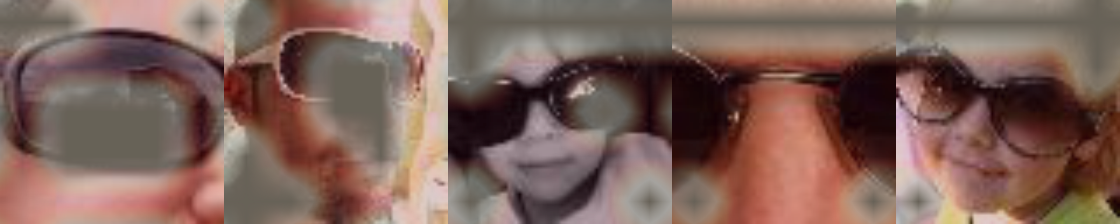}}
\end{minipage}\hrule
\vspace{3mm}
\medskip

\caption{Global explanations of ResNet50 for random trials at $c^\prime = 32$, highlighting higher prototype consistency compared to fidelity. Each class's prototypes correspond to the most significant of the 32 prototypes in each scenario.}
\vspace{-1cm}
\label{more_random}
\end{figure}

As shown in Fig. \ref{best_faith_plot}, the CAV predictions (blue dots) compared to the ground truth indicate higher Coherence and \textit{Faith} scores, outperforming Fidelity: \textit{Faith} scores of $96.9 \pm 3\%$ ($99.9\%$ Coherence, $94\%$ Fidelity) and $98.7 \pm 1\%$ ($99.9\%$ Coherence, $97\%$ Fidelity) for ResNet50, tested on an image of an Australian Kelpie and a Chihuahuaa respectively, and $98.5 \pm 1\%$ ($99.5\%$ Coherence, $97.5\%$ Fidelity) and $96.7 \pm 2$ ($98.8\%$ Coherence, $94.6\%$ Fidelity) for the InceptionV3, tested on an image of a Macaw and an Eagle respectively. The \textit{Faith} score is the mean value between Coherence (upper bound) and Fidelity (lower bound). Our extensive experiments on several classes show that Coherence scores consistently exceed Fidelity scores, revealing that concepts are more consistent than accurate (see Fig.~\ref{more_random}). Fidelity inversely relates to the sparsity of CAV predictions relative to the ground truth: increased dispersion reduces fidelity, while closer alignment increases it. High coherence, however, is directly linked to the positive correlation between CAV predictions and the ground truth, regardless of their distribution's sparsity. Low dispersion and high correlation of CAV predictions indicate that concept explanations are both accurate and consistent, validating meaningful explanations \cite{ace}. Combining Fidelity and Coherence scores balances their weaknesses, providing a more reliable metric for comprehensive quantitative assessment.

\begin{table*}[!t]
 \centering
\caption{Faithfulness comparison of explanation methods for ResNET50 at $c^\prime$ = 32.}
\label{compare_concepts_resnet}
\begin{tabular}{P{10pt}P{70pt}P{18pt}P{27pt}P{17pt}|P{18pt}P{27pt}P{17pt}|P{18pt}P{27pt}P{17pt}|P{18pt}P{27pt}P{17pt}}
    \hline
     && \multicolumn{3}{c}{Australian Kelpie} & \multicolumn{3}{c}{Chihuahua} & \multicolumn{3}{c}{Macaw} & \multicolumn{3}{c}{Eagle}\\
    \hline
    & Explainers & Fidelity & Coherence & \textit{Faith} & Fidelity & Coherence & \textit{Faith}& Fidelity & Coherence & \textit{Faith}& Fidelity & Coherence & \textit{Faith}\\
    \hline
  \multirow{12}{*}{\rotatebox[origin=c]{90}{\makecell{Attribution \\ based}}} 
  & GradCAM\cite{gradcam}  & 90.8\%  & 28.6\% & 59.7\% & 86.5\% & 26.5\% & 56.5\% & 94.3\%  & 40.7\% & 67.5\% & 94.2\% & 34.5\% & 64.4\%\\
  & GradCAM++ \cite{gradcanplusplus}  & 85.2\%  & 28.6\% & 56.9\% & 82.0\% & 27.9\% & 54.9\% & 90.9\%  & 39.8\% & 65.3\% & 89.2\% & 35.0\% & 62.1\%\\
  & HiResCAM \cite{hirescam} & 90.8\% & 28.6\% & 59.7\% & 86.5\% & 26.5\% & 56.5\% & 94.3\%  & 40.7\% & 67.5\% & 94.2\% & 34.5\% & 64.4\%\\
  & FullGrad \cite{fullgrad}  &  74.1\% & 28.7\% & 51.4\% & 73.0\% & 29.0\% & 54.0\% & 81.1\%  & 36.7\% & 58.9\% & 79.7\% & 35.4\% & 57.6\%\\
  & AblationCAM \cite{ablationcam} &  86.5\% & 28.5\% & 57.5\% & 79.1\% & 27.8\% & 53.4\% & 92.2\%  & 39.9\% & 66.1\% & 88.9\% & 35.7\% & 62.3\%\\
  & EigenCAM \cite{eigencam}  & 56.6\%  & 23.4\% & 40.0\% & 56.6\% & 23.4\% & 40.0\% & 89.9\%  & 37.7\% & 63.8\% & 89.5\% & 37.7\% & 63.8\%\\
  & EigenGradCAM \cite{eigencam} & 53.2\% & 23.7\% & 38.4\% & 49.4\% & 23.9\% & 36.6\% & 45.0\% & 29.4\% & 37.2\% & 45.0\% & 29.4\% & 37.2\%\\
  & LayerCAM \cite{layercam} & 83.9\% & 28.4\% & 56.1\% & 78.3\% & 27.6\% & 52.9\% & 89.4\% & 38.2\% & 63.8\% & 61.4\% & 29.4\% & 37.2\%\\

  & LIME\cite{why_trust_you}  & 80.4\%  & 88.8\% & 84.6\% & 82.5\% & 79.8\% & 81.2\% & 81.1\%  & 88.8\% & 85.0\% & 80.3\% & 88.8\% & 84.5\% \\
  & SHAP\cite{why_trust_you}  & 82.0\% & 27.9\% & 54.9\% & 90.3\% & 98.8\% & 94.5\% & 90.4\%  & 98.8\% & 94.6\% & 92.5\% & 99.1\% & 95.8\%\\
  & IG\cite{integrated_gradients}  & 62.3\%  & 76.9\% & 69.6\% & 72.9\% & 86.7\% & 79.8\% & 62.3\%  & 76.9\% & 69.6\% & 72.9\% & 82.3\% & 77.6\%\\
& DeepLift\cite{deeplift}  & 56.7\%  & 89.1\% & 72.9\%  & 92.1\%  & 82.3\% & 87.3\% & 86.5\% & 28.5\% & 57.5\% & 79.1\% & 27.8\% & 53.4\%\\

& GradientSHAP\cite{gradientshap}  & 85.8\% & 80.4\%  & 82.9\% & 56.7\% & 89.1\% & 72.9\%  &  94.2\% & 86.5\% & 90.3\% & 62.3\% & 77.0\% & 69.6\%   \\

  \hline
  \multirow{4}{*}{\rotatebox[origin=c]{90}{\makecell{Concept \\ based}}}
  & ACE\cite{ace}  & 91.1\%  & 98.8\% & 95.0\% & 90.3\% & 98.8\% & 94.5\% & 90.4\%  & 98.8\% & 94.6\% & 92.5\% & 99.1\% & 95.8\%\\
  & ICE (NMF)\cite{ice}  & 95.2\% & 99.6\% & 97.4\% & 94.3\% & \underline{99.5\%} & 96.9\% & 95.2\%  & 99.7\% & 97.4\% & 96.8\% & \underline{99.8\%} & 98.3\%\\
    & ICE (PCA) \cite{ice}  & 97.4\% & \underline{99.8\%} & 98.5\% & 96.6\% & \textbf{99.8\%} & 98.2\% & 98.0\%  & \underline{99.8\%} & 98.8\% & \textbf{98.7\%} & \underline{99.8\%} & \textbf{99.3\%} \\
  & CRAFT \cite{craft} & \underline{97.6\%} & \underline{99.8\%} & \underline{98.7\%} & \underline{98.2}\% & \textbf{99.8\%} & \underline{99.0\%} & \textbf{98.8\%}  & \underline{99.8\%} & \textbf{99.3\%} & \underline{98.6\%} & 99.4\% & 99.0\%\\
  & TraNCE (ours)  &  \textbf{97.9\%} & \textbf{99.9\%} & \textbf{98.9\%} & \textbf{98.4\%} & \textbf{99.8\%} & \textbf{99.1\%} & \underline{98.3\%}  & \textbf{99.9\%} & \underline{99.1\%} & 98.2\% & \textbf{100.0\% } & \underline{99.1\%} \\  
  \hline
\end{tabular}
\end{table*}
\begin{table*}[!t]
 \centering
\caption{Faithfulness comparison of explanation methods for InceptionV3 at $c^\prime$ = 32.}
\label{compare_concepts_incpetion}
\begin{tabular}{P{10pt}P{70pt}P{18pt}P{27pt}P{17pt}|P{18pt}P{27pt}P{17pt}|P{18pt}P{27pt}P{17pt}|P{18pt}P{27pt}P{17pt}}
    \hline
     && \multicolumn{3}{c}{Australian Kelpie} & \multicolumn{3}{c}{Chihuahua} & \multicolumn{3}{c}{Macaw} & \multicolumn{3}{c}{Eagle}\\
    \hline
    & Explainers & Fidelity & Coherence & \textit{Faith} & Fidelity & Coherence & \textit{Faith}& Fidelity & Coherence & \textit{Faith}& Fidelity & Coherence & \textit{Faith}\\
    \hline
  \multirow{12}{*}{\rotatebox[origin=c]{90}{\makecell{Attribution \\ based}}} 
  & GradCAM\cite{gradcam}  & 90.3\%  & 76.1\% & 83.2\% & 94.6\% & 91.0\% & 92.8\% & 94.2\%  & 84.0\% & 89.1\% & 94.2\% & 78.6\% & 86.4\%\\
  & GradCAM++ \cite{gradcanplusplus}  & 91.2\%  & 89.4\% & 90.3\% & 83.1\% & 78.6\% & 80.9\% & 87.1\%  & 82.9\% & 85.0\% & 82.5\% & 79.8\% & 81.2\%\\
  & HiResCAM \cite{hirescam} & 90.2\% & 76.1\% & 83.2\% & 94.6\% & 91.0\% & 92.8\% & 94.2\%  & 78.6\% & 86.3\% & 94.2\% & 84.0\% & 89.1\%\\
  & FullGrad \cite{fullgrad}  &  -- & -- & -- & -- & -- & --& --  & -- & -- & -- & -- & --\\
  & AblationCAM \cite{ablationcam} &  94.2\% & 86.5\% & 90.3\% & 92.1\%  & 82.3\% & 87.3\% & 86.5\% & 28.5\% & 57.5\% & 79.1\% & 27.8\% & 53.4\%\\
  & EigenCAM \cite{eigencam}  & 56.7\%  & 89.1\% & 72.9\% & 56.7\% & 89.1\% & 72.9\% & 62.3\% & 77.0\% & 69.6\% & 62.3\%  & 76.9\% & 69.6\%\\
  & EigenGradCAM \cite{eigencam} & 92.2\%  & 76.0\% & 84.1\% & 47.2\% & 88.2\% & 67.7\% & 94.2\% & 86.5\% & 90.3\% & 92.1\%  & 82.3\% & 87.3\%\\
  & LayerCAM \cite{layercam} & 80.0\% & 80.0\% & 80.0\% & 88.6\% & 86.8\% & 87.7\% & 89.4\% & 38.2\% & 63.8\% & 81.0\% & 80.0\% & 80.5\%\\

    & LIME\cite{why_trust_you}  & 79.3\%  & 89.4\% & 84.35\% & 81.3\% & 95.8\% & 88.6\% & 87.4\%  & 91.8\% & 89.6\% & 84.5\% & 91.1\% & 87.8\%\\
  & SHAP\cite{why_trust_you}  & 92.1\%  & 82.3\% & 87.3\% &  47.2\% & 88.2\% & 67.7\% & 80.4\%  & 88.8\% & 84.6\% & 82.5\% & 89.1\% & 85.8\%\\
  & IG\cite{integrated_gradients}  & 80.4\%  & 88.8\% & 84.6\% & 82.5\% & 89.1\% & 95.8\% & 81.1\%  & 88.8\% & 85.0\% & 80.3\% & 88.8\% & 84.5\% \\
& DeepLift\cite{deeplift}  & 72.9\% & 86.7\% & 79.8\% & 85.8\% & 80.4\%  & 82.9\% & 62.3\%  & 76.9\% & 69.6\% & 72.9\% & 82.3\% & 77.6\%\\

& GradientSHAP\cite{gradientshap}  & 82.5\% & 89.1\% & 85.8\% & 80.4\%  & 88.8\% & 80.4\%  & 80.3\% & 88.8\% & 84.5\% & 88.8\% & 84.6\% & 95.8\%\\

  \hline
  \multirow{4}{*}{\rotatebox[origin=c]{90}{\makecell{Concept \\ based}}}
  & ACE\cite{ace}  & 90.6\%  & 98.8\% & 94.7\% & 86.3\% & 98.0\% & 92.1\% & 87.8\%  & 98.3\% & 93.0\% & 94.3\% & 99.4\% & 96.8\%\\
  & ICE (NMF) \cite{ice}  & 94.7\% & 99.5\% & 97.1\% & 93.6\% & 99.5\% & 96.5\% & 93.3\%  & 99.4\% & 96.4\% & 98.0\% & \textbf{100.0\%} & 99.0\%\\
  & ICE (PCA) \cite{ice}  & 96.3\% & \underline{99.8\%} & 98.0\% & 96.3\% & \underline{99.8\%} & 98.1\% & \underline{98.0\%}  & \textbf{100.0\%} & \underline{99.0\%} & \underline{98.7\%} & \textbf{100.0\%} & \underline{99.3\%} \\
  & CRAFT \cite{craft} & \textbf{96.7\%} & \textbf{99.9\%} & \textbf{98.3\%} & \underline{98.4\%} & \underline{99.8\%} &\underline{99.1\%}  & \underline{98.0\%} & 99.6\% & 98.8\% & \underline{98.7\%} & \underline{99.7\%} & 99.2\%\\
  & TraNCE (ours)  &  \underline{96.6\%} & \underline{99.8\%} & \underline{98.2\%} & \textbf{98.7\%} & \textbf{99.9\%} & \textbf{99.3\%} & \textbf{98.5\%}  & \underline{99.9\%} & \textbf{99.2\%} & \textbf{99.2\%} & \textbf{100.0\% } & \textbf{99.6\%}\\  
  \hline
\end{tabular}
\end{table*}

\subsection{Comparison with SOTA methods}

We compared the quality of concept explanations using our proposed \textit{Faith} scoring criteria, conducting head-to-head comparisons of the proposed TraNCE against ICE \cite{ice}, ACE \cite{ace}, CRAFT \cite{craft} and attribution methods \cite{gradcam,gradcanplusplus,hirescam,fullgrad,ablationcam,eigencam,layercam}. While comparing TraNCE with attribution methods may be futile since they only provide qualitative explanations, it offers more insights into the benefit of concept-based methods. Furthermore, it demonstrates the benefit of the proposed \textit{Faith} scoring metric for faithfulness evaluation, offering an opportunity to introduce quantitative evaluation of the attribution methods. We computed \textit{Faith} scores for each attribution-based method by comparing the respective methods' activation and the CNN's logits as ground truth. Results are shown in Tables \ref{compare_concepts_resnet} and \ref{compare_concepts_incpetion}. The highest scores are highlighted in bold while the penultimate scores are underlined.

The results in Tables \ref{compare_concepts_resnet} and \ref{compare_concepts_incpetion} and Fig.\ref{more_random} demonstrate the benefit of TraNCE, which achieves much higher scores than all of the attribution methods and marginally outperforms the concept-based methods. TraNCE yields notably higher \textit{Faith} scores, showing that it discovers more consistently accurate concepts from the respective models' feature maps. We also observed that for an arbitrary model (ResNet50 for instance), any explainer (including the proposed TraNCE) could produce highly prototypical concepts for an arbitrary target class (Chihuahua for instance) while also performing poorly for another target class. This does not reflect a limitation in the explainer, but rather the CNN's classification inefficiencies for such a target class. In simple terms, a high true positive rate for an arbitrary target class would imply that the discovered concepts for the class are highly prototypical, and vice versa for an arbitrary target class with a high false positive rate.

Unlike other methods, the flexible/adaptive architecture of TraNCE offers a potential for broader utility, despite concerns about computational costs. Such an adaptive architecture further makes TraNCE more suited for explanation cases where the complexity of relationships in the feature map exceeds the capabilities of linear reducers, offering increased utility for explanation tasks for FGVC tasks with small inter-class variance. 

\subsection{Performance on a varying number of user-defined concepts and at different intermediate layers}

\begin{figure*}[!t]
\begin{minipage}[b]{1\linewidth}
\centering\centerline{\includegraphics[width=1\linewidth]{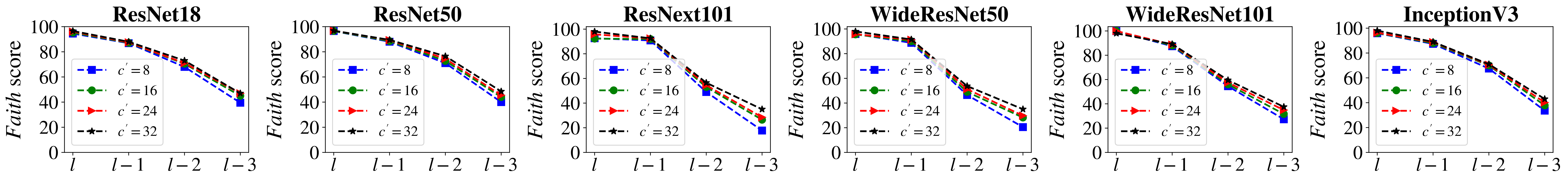}}
  \centerline{(a)}\medskip
\end{minipage}

\begin{minipage}[b]{1\linewidth}
\centering\centerline{\includegraphics[width=1\linewidth]{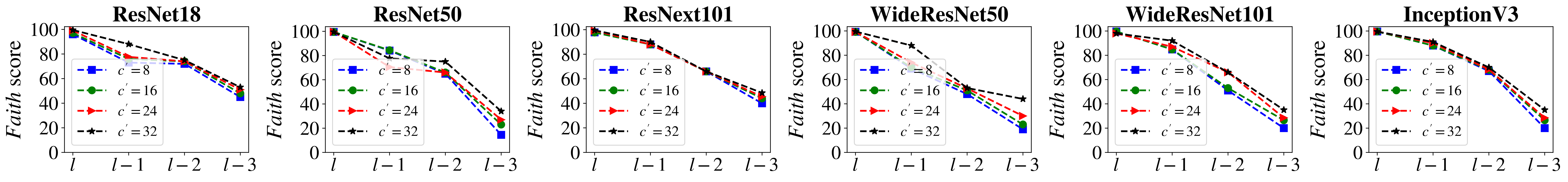}}
  \centerline{(b)}\medskip
\end{minipage}

\begin{minipage}[b]{1\linewidth}
\centering\centerline{\includegraphics[width=1\linewidth]{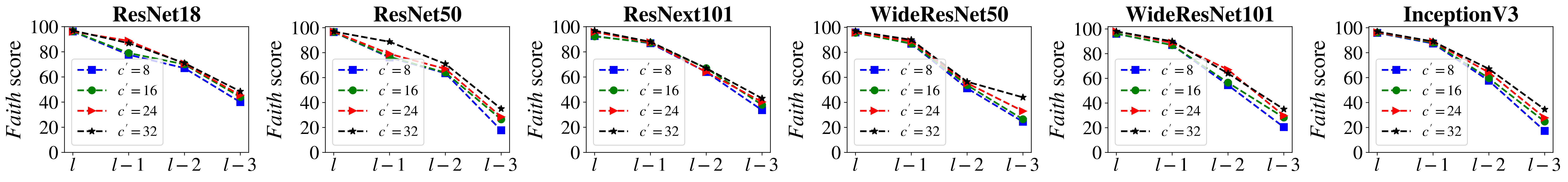}}
  \centerline{(c)}\medskip
\end{minipage}

\begin{minipage}[b]{1\linewidth}
\centering\centerline{\includegraphics[width=1\linewidth]{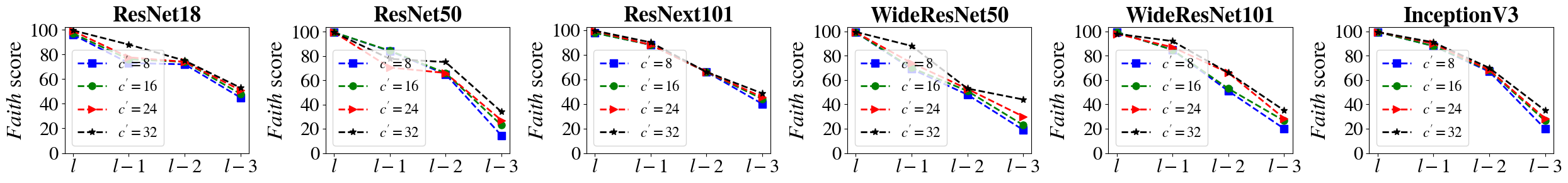}}
  \centerline{(d)}\medskip
\end{minipage}
\vspace*{-5mm}
\caption{Impact of increasing $c^\prime$ and at different CNN layers (a) ResNet50's explanation for Australian Kelpie, (b) ResNet50's explanation for Chihuahua, (c) InceptionV3's explanation for Macaw, and (d) InceptionV3's explanation for Eagle.}
\label{diff_concepts}
\end{figure*}

CNNs exhibit the capacity to generate increasingly refined and discriminative features at deeper layers, as indicated by \cite{craft}, suggesting their potential applicability to a diverse range of conceptual understanding in visual data processing tasks. To verify this, we explored the impact of user-defined concept numbers ($c^\prime = {8, 16, 24, 32}$) on \textit{Faith} scores across various CNN intermediate layers ($k = {l, l-1, l-2, l-3}$). Fig. \ref{diff_concepts} shows the impact of increasing $c^\prime$ and at different CNN intermediate layers on \textit{Faith} scores of pretrained models. 

The decreasing trend in the figures shows that while the explanations are faithful at the deepest intermediate layer ($k = l$) for all the CNN models, increasing the number of concepts only marginally increases explainer faithfulness. However, targeting shallow layers results in a decrease in explainer faithfulness. This reveals that the quality of features produced at deeper layers is better. These features are more discriminative and contain local features that the explainer harnesses to produce quality concepts.

\section{Discussion}
\label{discussion}
\subsection{Sanity Checks}
\label{sanity}
\begin{figure}[!t]
\centering
    \begin{subfigure}[c]{1\linewidth}
        \centering 
        \hspace*{3mm}\leftline{ \textbf{Test} }\vspace*{-4mm} 
        \hspace*{16mm}\leftline{ \textbf{Concept} }\vspace*{-4mm}
        \hspace*{-55mm}\rightline {\textbf{Prototypes}}\vspace*{1mm}\\
        \hrule \vspace*{1mm}
    \end{subfigure}
\begin{subfigure}[tc]{0.18\linewidth}
        \raisebox{-\height}{\includegraphics[width=\linewidth]{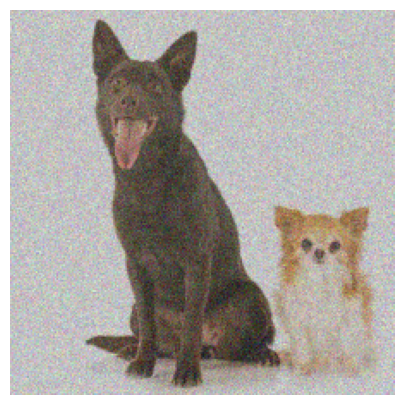}} \vspace*{-1mm}\\
        \centerline{\footnotesize ($\sigma = 0.1$)}
    \end{subfigure}
    \begin{subfigure}[c]{0.8\linewidth}
        \centering
        \raisebox{-\height}{\includegraphics[width=0.165\linewidth]{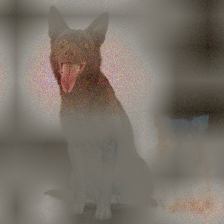}}\hspace*{0.5mm} \raisebox{-\height}{\includegraphics[width=0.82\linewidth]{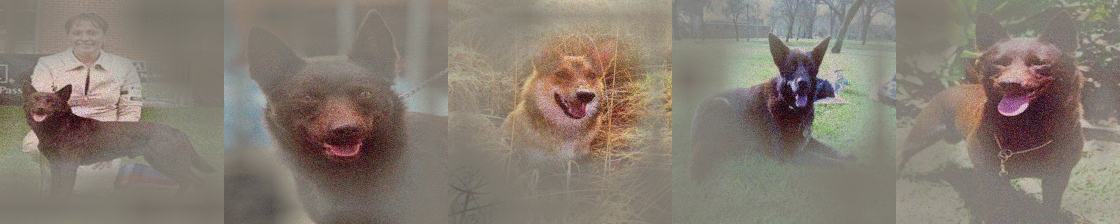}}\vspace*{0.7mm}\\
        \raisebox{-\height}{\leftline{\footnotesize Kelpie: $\mathcal{\zeta}= 0.82$, \textit{Faith}= $98.6\pm1.4\%$, Similarity = 0.71.}}\\\vspace*{-1.2mm}
       
        \raisebox{-\height}{\includegraphics[width=0.165\linewidth]{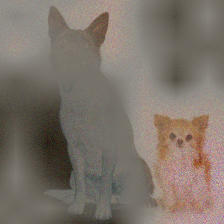}}\hspace*{0.5mm} \raisebox{-\height}{\includegraphics[width=0.82\linewidth]{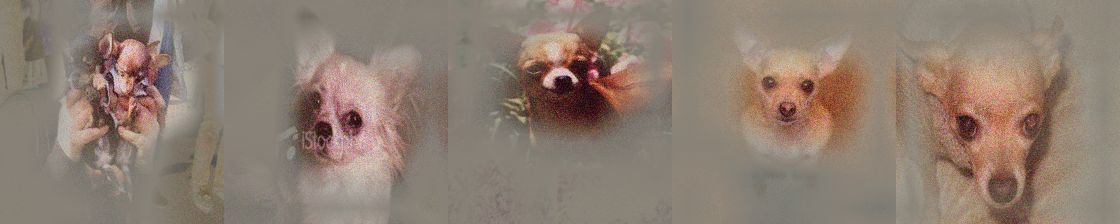}}\vspace*{0.7mm}\\
         \raisebox{-\height}{\leftline{\footnotesize Chihuahua:$\mathcal{\zeta}=0.79$, \textit{Faith} = $98.4\pm1.5\%$, Similarity = 0.62}}\\
    \end{subfigure}
     \vspace*{1mm} \hrule \vspace*{1mm}
    \begin{subfigure}[tc]{0.18\linewidth}
        \raisebox{-\height}{\includegraphics[width=\linewidth]{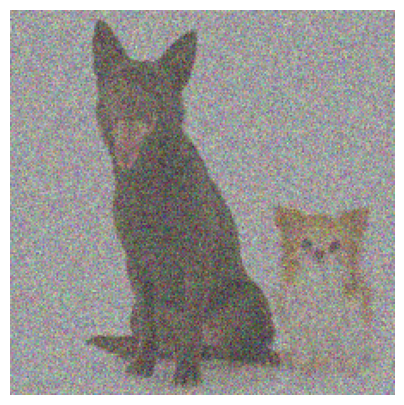}} \vspace*{-1mm}\\
        \centerline{\footnotesize ($\sigma = 0.3$)}
    \end{subfigure}
    \begin{subfigure}[c]{0.8\linewidth}
        \centering
        \raisebox{-\height}{\includegraphics[width=0.165\linewidth]{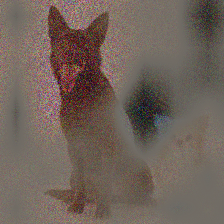}}\hspace*{0.5mm} \raisebox{-\height}{\includegraphics[width=0.82\linewidth]{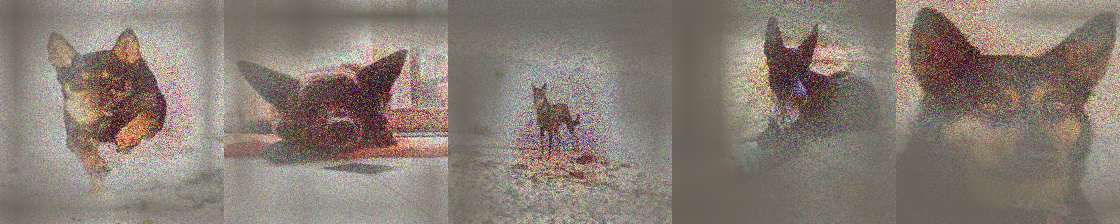}}\hspace{0.25\linewidth}\vspace*{-4mm}\\
        \raisebox{-\height}{\leftline{\footnotesize Kelpie: $\mathcal{\zeta}= 0.51$, \textit{Faith}= $94.9\pm 4.6\%$, Similarity = 0.43}}\\\vspace*{-1.2mm}
        \raisebox{-\height}{\includegraphics[width=0.165\linewidth]{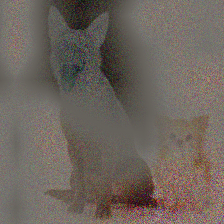}}\hspace*{0.5mm} \raisebox{-\height}{\includegraphics[width=0.82\linewidth]{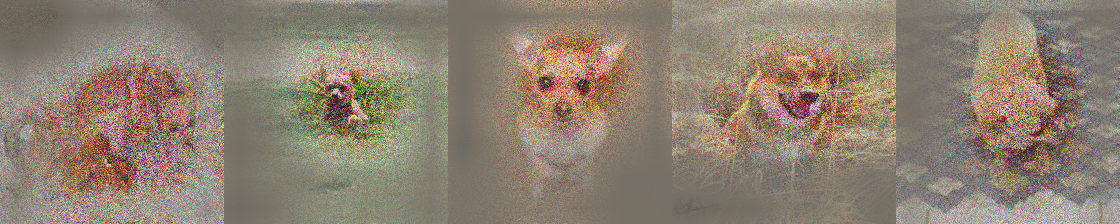}}\vspace*{0.7mm}\\
        \raisebox{-\height}{\leftline{\footnotesize Chihuahua: $\mathcal{\zeta}= 0.47$, \textit{Faith}= $89.8\pm9.0\%$, Similarity = 0.31}}\\
    \end{subfigure}
\vspace*{1mm} \hrule \vspace*{1mm}
    \begin{subfigure}[tc]{0.18\linewidth}
        \raisebox{-\height}{\includegraphics[width=\linewidth]{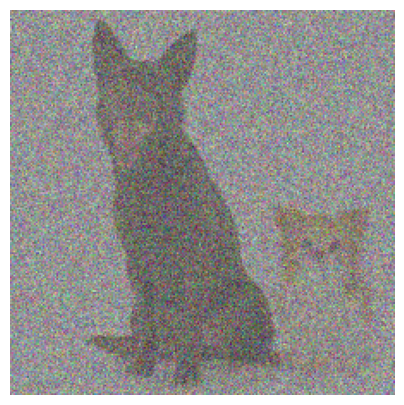}} \vspace*{-1mm}\\
        \centerline{\footnotesize ($\sigma = 0.5$)}
    \end{subfigure}
    \begin{subfigure}[c]{0.8\linewidth}
        \centering
        \raisebox{-\height}{\includegraphics[width=0.165\linewidth]{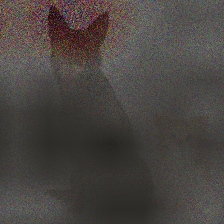}}\hspace*{0.5mm} \raisebox{-\height}{\includegraphics[width=0.82\linewidth]{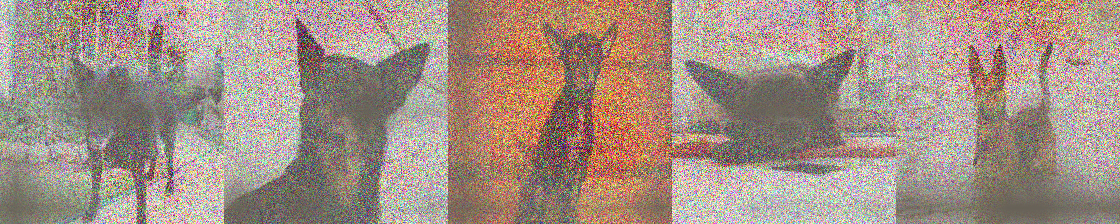}}\hspace{0.25\linewidth}\vspace*{-4mm}\\
        \raisebox{-\height}{\leftline{\footnotesize Kelpie: $\mathcal{\zeta}= 0.14$, \textit{Faith}= $56.9\pm28\%$, Similarity = 0.08}}\\\vspace*{-1.2mm}
        \raisebox{-\height}{\includegraphics[width=0.165\linewidth]{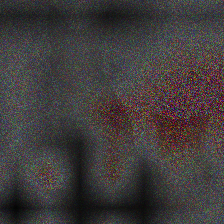}}\hspace*{0.5mm} \raisebox{-\height}{\includegraphics[width=0.82\linewidth]{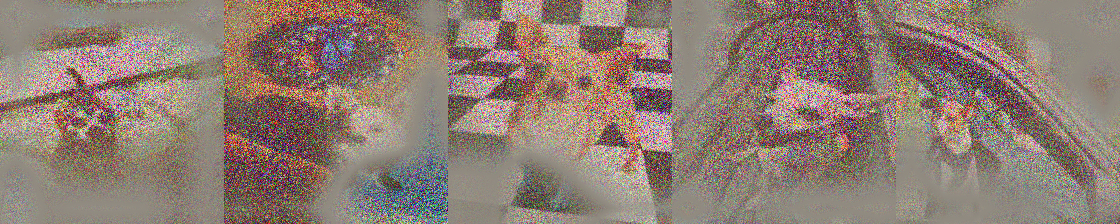}}\vspace*{0.7mm}\\
        \raisebox{-\height}{\leftline{\footnotesize Chihuahua: $\mathcal{\zeta}= 0.11$, \textit{Faith}= $53.4 \pm 31\%$, Similarity = 0.09}}\\
    \end{subfigure}
\vspace*{1mm} \hrule \vspace*{3mm}
    \hfill
\centerline{(a)}\medskip\vspace*{2mm}
    \begin{minipage}[b]{.49\linewidth}
      \centerline{\includegraphics[width=1\linewidth]{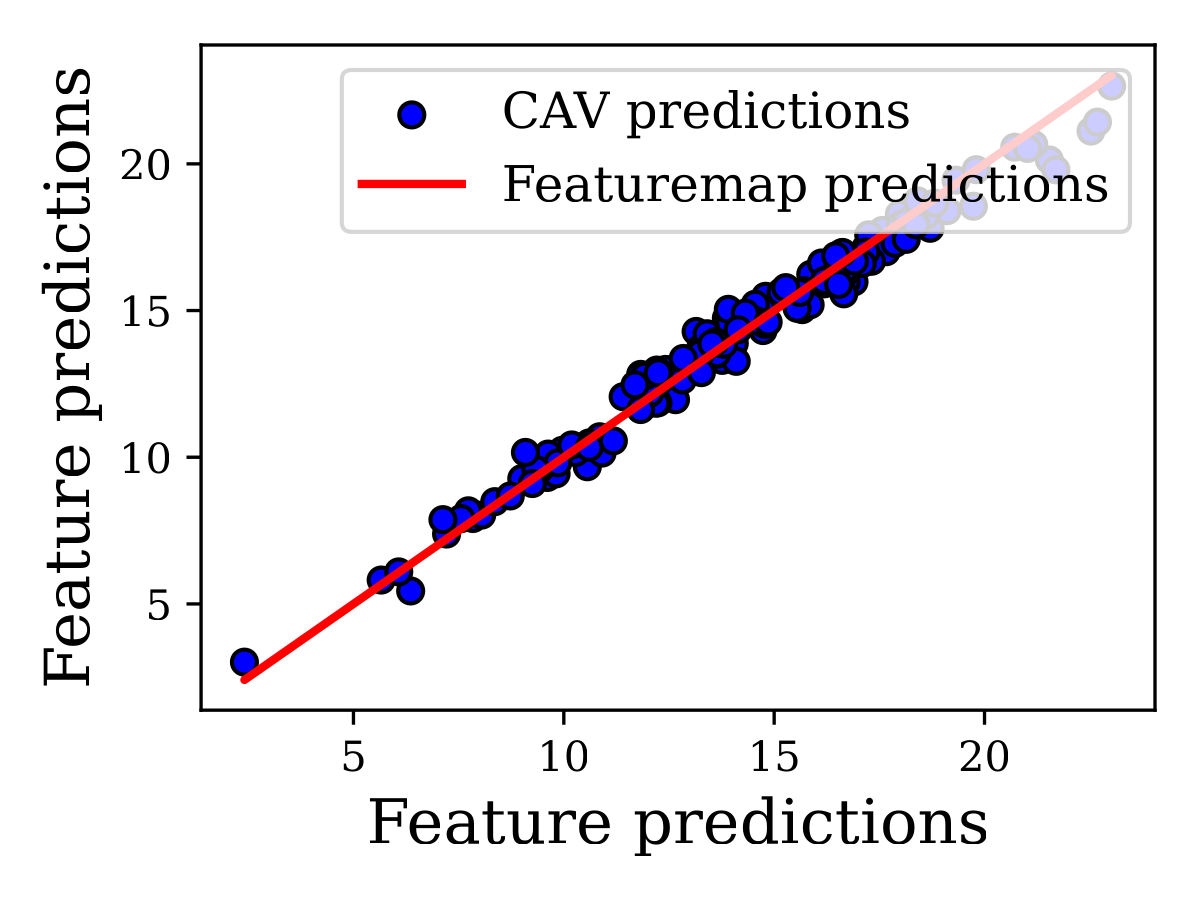}}
      \vspace*{-12ex}  \hspace*{2ex}
        \begin{center}
       {\hspace*{8ex} \small Kelpie: $\sigma = 0.1$  }
        \end{center}\vspace*{3mm}\medskip
        \vspace*{12ex} \hspace*{.2ex} 
    \end{minipage} 
    \begin{minipage}[b]{.49\linewidth}
      \centerline{\includegraphics[width=1\linewidth]{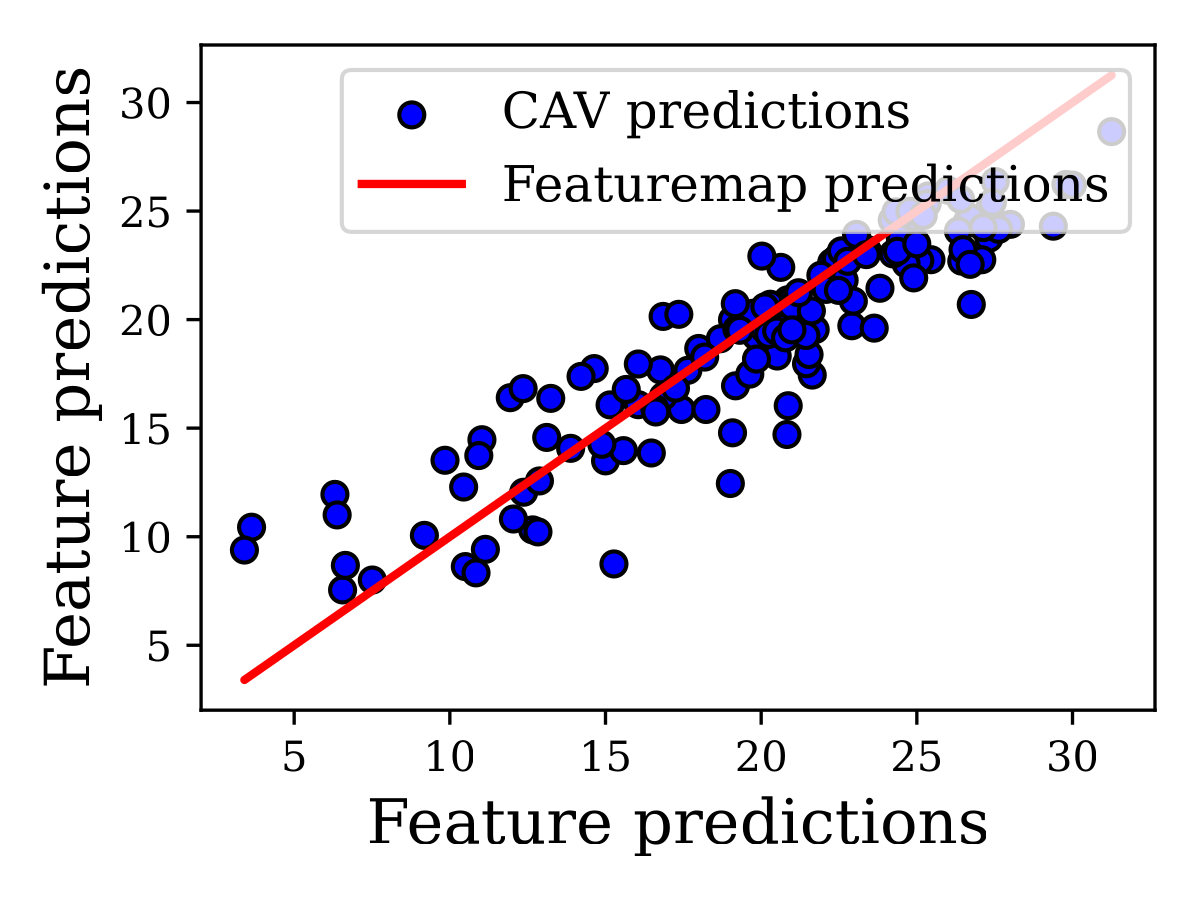}}
      \vspace*{-12ex} \hspace*{-.2ex}
        \begin{center}
        {\hspace*{7ex} \small Kelpie: $\sigma = 0.5$  }
        \end{center}\vspace*{3mm} \medskip
        \vspace*{12ex} \hspace*{.2ex} 
    \end{minipage}\vspace*{-22mm}\vspace*{1mm}
\centerline{(b)}
\caption{Top ResNet50 explanations for an Australian Kelpie and a Chihuahua (a) with added Gaussian noise ($\sigma = \{0.1, 0.3, 0.5\}$) and (b) faithfulness at $\sigma = 0.1$ and $0.5$. }
\label{noisy}
\end{figure}

\begin{figure}[!t]
\centering
    \begin{subfigure}[c]{1\linewidth}
        \centering 
        \hspace*{3mm}\leftline{ \textbf{Test} }\vspace*{-4mm} 
        \hspace*{16mm}\leftline{ \textbf{Concept} }\vspace*{-4mm}
        \hspace*{-55mm}\rightline {\textbf{Prototypes}}\vspace*{1mm}\\
        \hrule \vspace*{1mm}
    \end{subfigure}

\begin{subfigure}[tc]{0.18\linewidth}
        \raisebox{-\height}{\includegraphics[width=\linewidth]{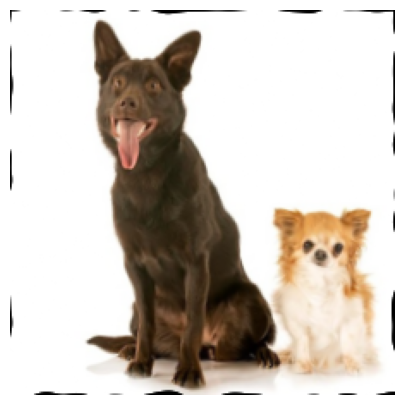}} \vspace*{-1mm}\\
        \centerline{\footnotesize ($\alpha = 50.0$)}
    \end{subfigure}
    \begin{subfigure}[c]{0.8\linewidth}
        \centering
        \raisebox{-\height}{\includegraphics[width=0.165\linewidth]{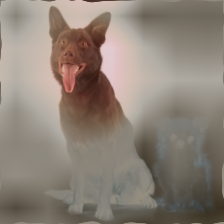}}\hspace*{0.5mm} \raisebox{-\height}{\includegraphics[width=0.82\linewidth]{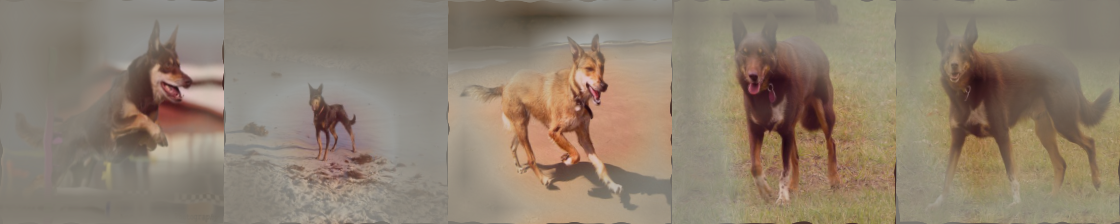}}\hspace{0.25\linewidth}\\\vspace*{0.7mm}
        \raisebox{-\height}{\leftline{\footnotesize Kelpie: $\mathcal{\zeta}= 0.63$, \textit{Faith}= $98.4\pm 1.5\%$, Similarity = 0.56.}}\\\vspace*{-1.2mm}
       
        \raisebox{-\height}{\includegraphics[width=0.165\linewidth]{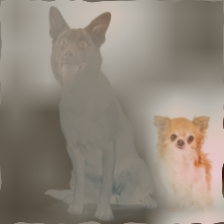}}\hspace*{0.5mm} \raisebox{-\height}{\includegraphics[width=0.82\linewidth]{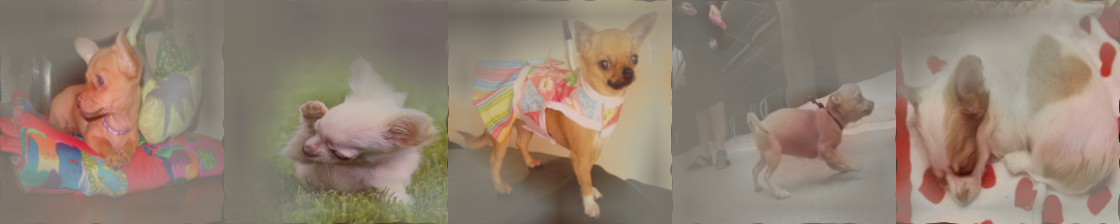}}\vspace*{0.7mm}\\
        \leftline{\footnotesize Chihuahua: $\mathcal{\zeta} = 0.81$, \textit{Faith} = $97.2\pm 2.7\%$, Similarity = 0.68.}
    \end{subfigure}
     \vspace*{1mm} \hrule \vspace*{1mm}
    \begin{subfigure}[tc]{0.18\linewidth}
        \raisebox{-\height}{\includegraphics[width=\linewidth]{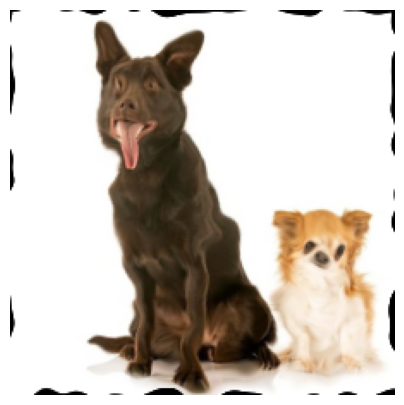}} \vspace*{-1mm}\\
        \centerline{\footnotesize ($\alpha = 100.0$)}
    \end{subfigure}
    \begin{subfigure}[c]{0.8\linewidth}
        \centering
        \raisebox{-\height}{\includegraphics[width=0.165\linewidth]{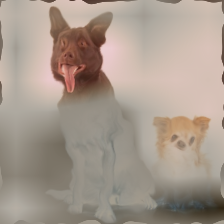}}\hspace*{0.5mm} \raisebox{-\height}{\includegraphics[width=0.82\linewidth]{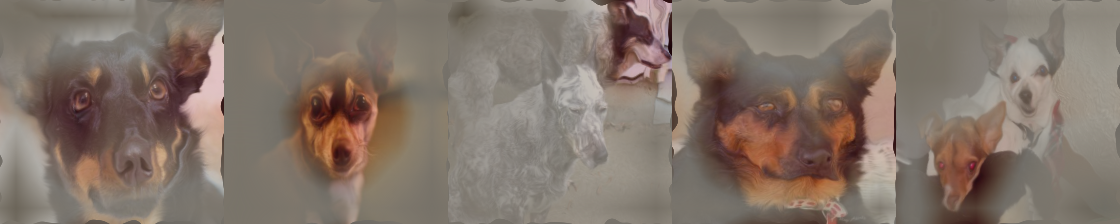}}\hspace{0.25\linewidth}\vspace*{-4mm}\\
        \raisebox{-\height}{\leftline{\footnotesize Kelpie: $\mathcal{\zeta}= 0.45$, \textit{Faith}= $96.5\pm 3.4\%$, Similarity = 0.31.}}\\\vspace*{-1.2mm}
        \raisebox{-\height}{\includegraphics[width=0.165\linewidth]{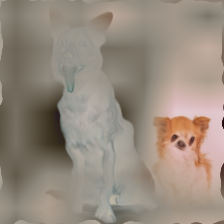}}\hspace*{0.5mm} \raisebox{-\height}{\includegraphics[width=0.82\linewidth]{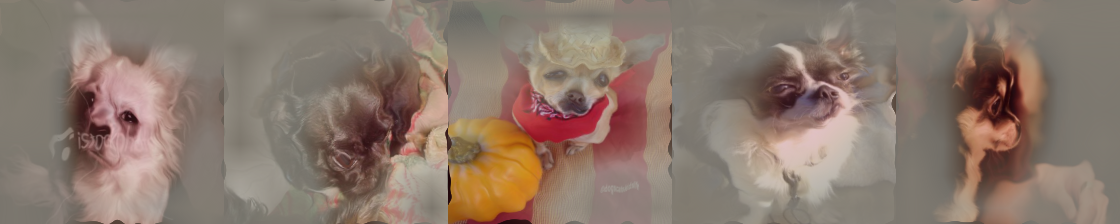}}\vspace*{0.7mm}\\
        \raisebox{-\height}{\leftline{\footnotesize Chihuahua: $\mathcal{\zeta}= 0.54$, \textit{Faith}= $97.4\pm2.5\%$, Similarity = 0.52.}}\\
    \end{subfigure}
\vspace*{1mm} \hrule \vspace*{1mm}
    \begin{subfigure}[tc]{0.18\linewidth}
        \raisebox{-\height}{\includegraphics[width=\linewidth]{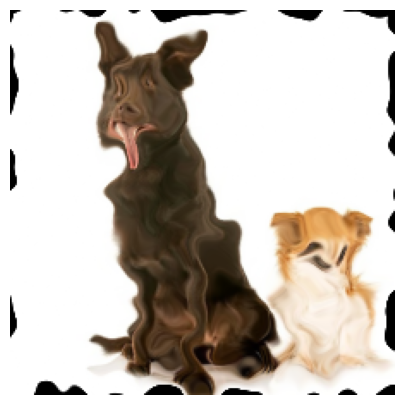}} \vspace*{-1mm}\\
        \centerline{\footnotesize ($\alpha = 200.0$)}
    \end{subfigure}
    \begin{subfigure}[c]{0.8\linewidth}
        \centering
        \raisebox{-\height}{\includegraphics[width=0.165\linewidth]{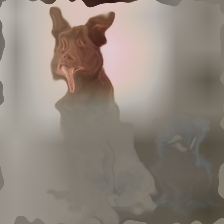}}\hspace*{0.5mm} \raisebox{-\height}{\includegraphics[width=0.82\linewidth]{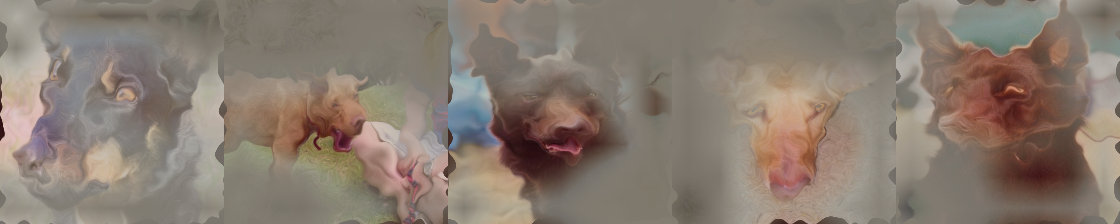}}\hspace{0.25\linewidth}\vspace*{-4mm}\\
        \raisebox{-\height}{\leftline{\footnotesize Kelpie: $\mathcal{\zeta}= 0.16$, \textit{Faith}= $95.6\pm5.1\%$, Similarity = 0.13.}}\\\vspace*{-1.2mm}
        \raisebox{-\height}{\includegraphics[width=0.165\linewidth]{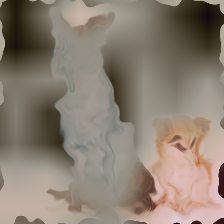}}\hspace*{0.5mm} \raisebox{-\height}{\includegraphics[width=0.82\linewidth]{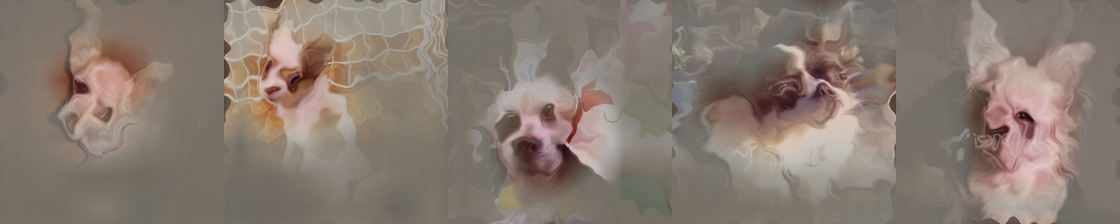}}\vspace*{0.7mm}\\
        \raisebox{-\height}{\leftline{\footnotesize Chihuahua: $\mathcal{\zeta}= 0.12$, \textit{Faith}= $91.5 \pm 7.1\%$, Similarity = 0.10.}}\\
    \end{subfigure}
    \vspace*{1mm} \hrule \vspace*{3mm}
    \hfill
\centerline{(a)}\medskip\vspace*{2mm}
    \begin{minipage}[b]{.49\linewidth}
      \centerline{\includegraphics[width=1\linewidth]{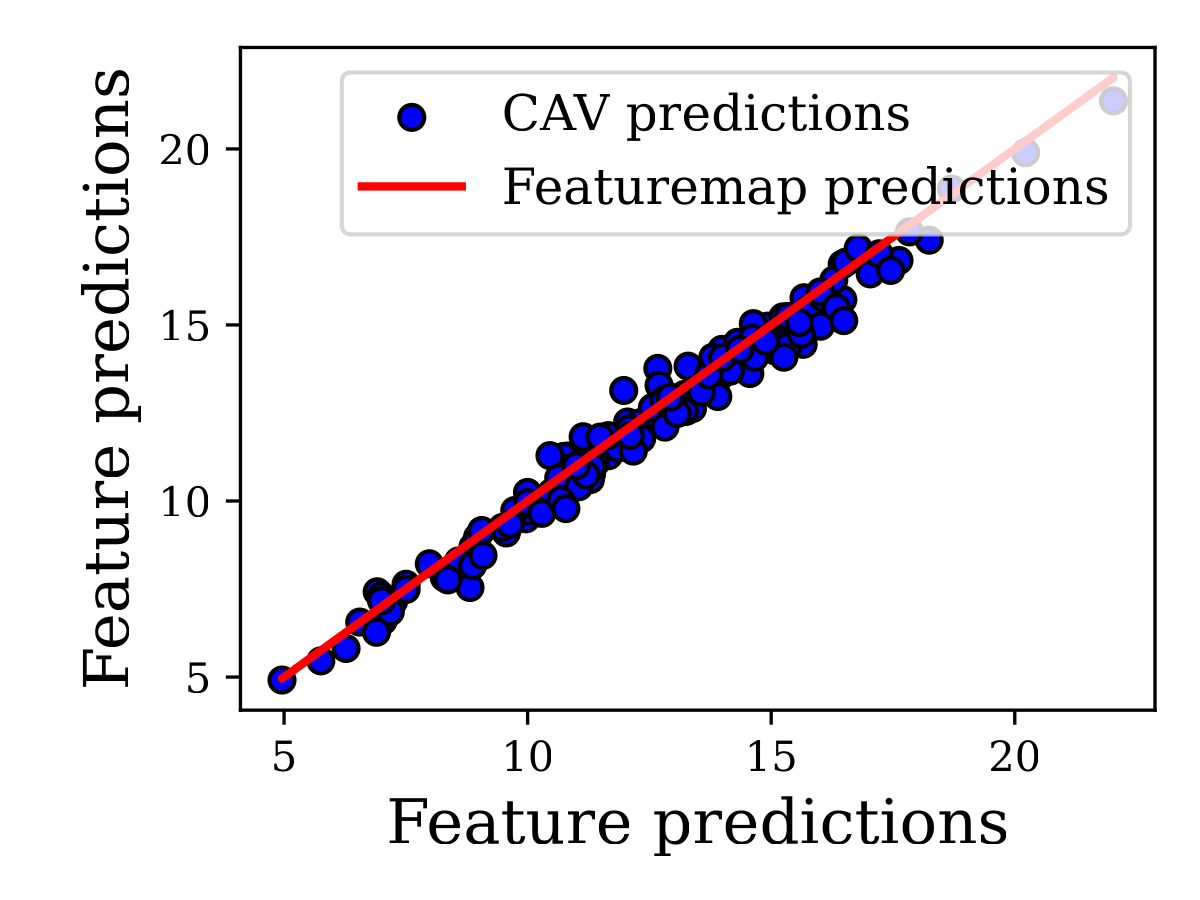}}
      \vspace*{-12ex}  \hspace*{2ex}  
        \begin{center}
       {\hspace*{8ex} \small Kelpie: $\alpha = 50.0$  }
        \end{center}\vspace*{3mm}\medskip
        \vspace*{12ex} \hspace*{.2ex} 
    \end{minipage} 
    \begin{minipage}[b]{.49\linewidth}
      \centerline{\includegraphics[width=1\linewidth]{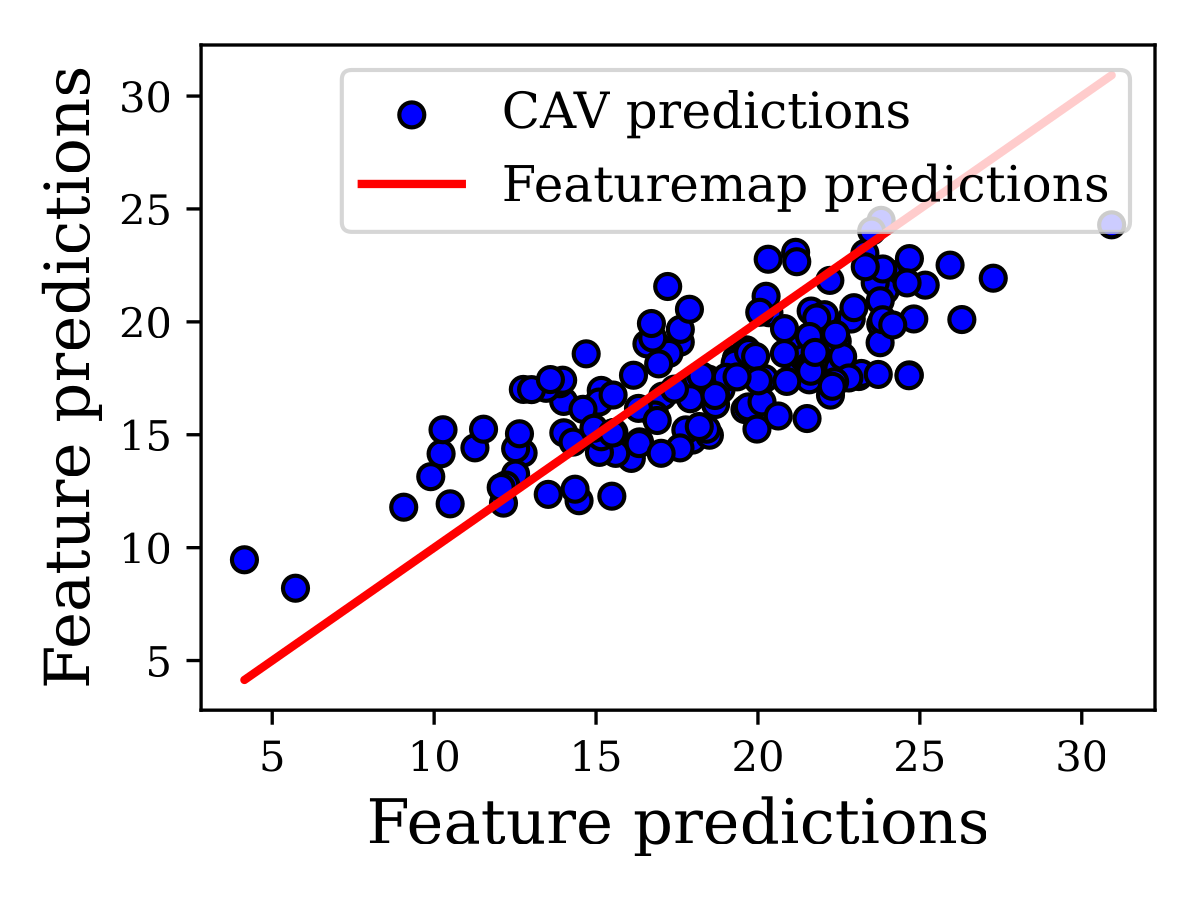}}
      \vspace*{-13ex} \hspace*{-.2ex}  
        \begin{center}
        {\hspace*{7ex} \small Kelpie: $\alpha = 200.0$  }
        \end{center}\vspace*{3mm} \medskip
        \vspace*{13ex} \hspace*{.2ex} 
    \end{minipage}\vspace*{-22mm}\vspace*{1mm}
\centerline{(b)}
\caption{Top ResNet50 explanations for an Australian Kelpie and a Chihuahua (a) with warping ($\alpha = \{50.0, 100.0, 200.0\}$, $\lambda = 0.5$) and (b) faithfulness at $\alpha = 50.0$ and $200.0$.}
\label{warp}
\end{figure}

\begin{figure}[!t]
\centering
    \begin{subfigure}[c]{1\linewidth}
        \centering 
        \hspace*{3mm}\leftline{ \textbf{Test} }\vspace*{-4mm} 
        \hspace*{16mm}\leftline{ \textbf{Concept} }\vspace*{-4mm}
        \hspace*{-55mm}\rightline {\textbf{Prototypes}}\vspace*{1mm}\\       
        \hrule \vspace*{1mm}
    \end{subfigure}
\begin{subfigure}[tc]{0.18\linewidth}
        \raisebox{-\height}{\includegraphics[width=\linewidth]{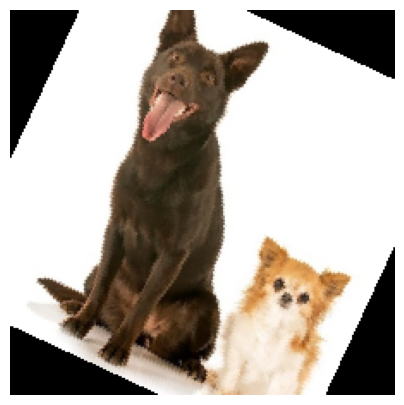}} \vspace*{-1mm}\\
        \centerline{\footnotesize ($\tau = 45\degree$)}
    \end{subfigure}
    \begin{subfigure}[c]{0.8\linewidth}
        \centering
        \raisebox{-\height}{\includegraphics[width=0.165\linewidth]{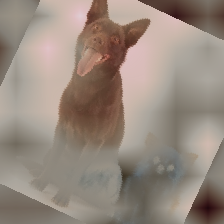}}\hspace*{0.5mm} \raisebox{-\height}{\includegraphics[width=0.82\linewidth]{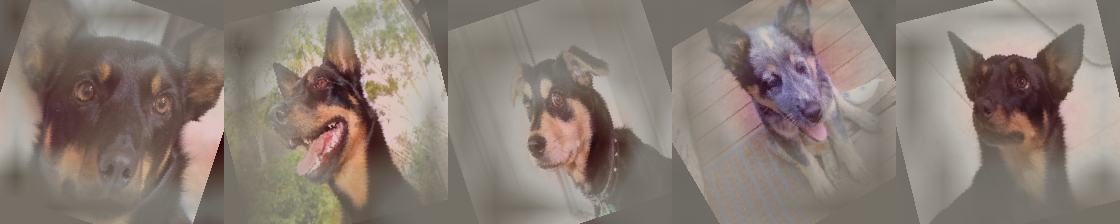}}\hspace{0.25\linewidth}\\\vspace*{0.7mm}
        \raisebox{-\height}{\leftline{\footnotesize Kelpie: $\mathcal{\zeta}= 0.71$, \textit{Faith}= $96.7\pm3.1\%$, Similarity = 0.61.}}\\\vspace*{-1.2mm}
       
        \raisebox{-\height}{\includegraphics[width=0.165\linewidth]{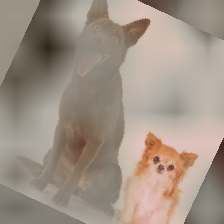}}\hspace*{0.5mm} \raisebox{-\height}{\includegraphics[width=0.82\linewidth]{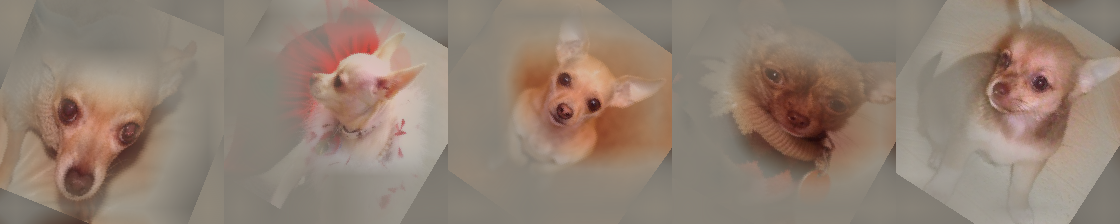}}\vspace*{0.7mm}\\
        \leftline{\footnotesize Chihuahua: $\mathcal{\zeta} = 0.72$, \textit{Faith} = $97.2\pm 2.5\%$, Similarity = 0.62.}
    \end{subfigure}
     \vspace*{1mm} \hrule \vspace*{1mm}
    \begin{subfigure}[tc]{0.18\linewidth}
        \raisebox{-\height}{\includegraphics[width=\linewidth]{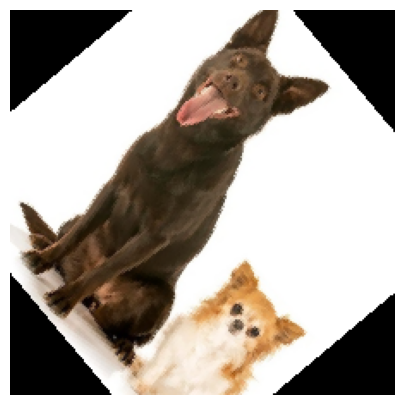}} \vspace*{-1mm}\\
        \centerline{\footnotesize ($\tau = 90\degree$)}
    \end{subfigure}
    \begin{subfigure}[c]{0.8\linewidth}
        \centering
        \raisebox{-\height}{\includegraphics[width=0.165\linewidth]{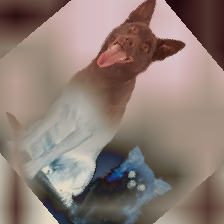}}\hspace*{0.5mm} \raisebox{-\height}{\includegraphics[width=0.82\linewidth]{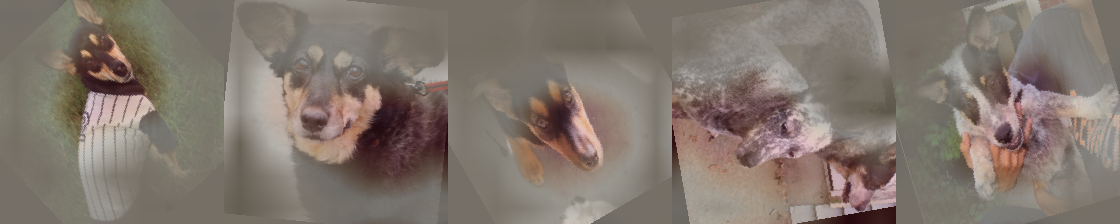}}\hspace{0.25\linewidth}\vspace*{-4mm}\\
        \raisebox{-\height}{\leftline{\footnotesize Kelpie: $\mathcal{\zeta}= 0.70$, \textit{Faith}= $97.0\pm 2.8\%$, Similarity = 0.63.}}\\\vspace*{-1.2mm}
        \raisebox{-\height}{\includegraphics[width=0.165\linewidth]{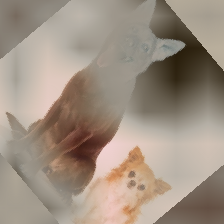}}\hspace*{0.5mm} \raisebox{-\height}{\includegraphics[width=0.82\linewidth]{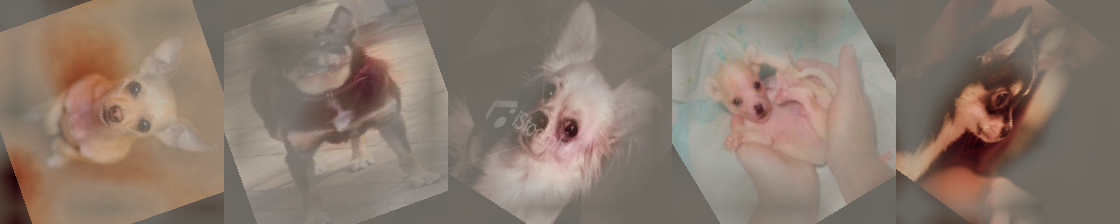}}\vspace*{0.7mm}\\
        \raisebox{-\height}{\leftline{\footnotesize Chihuahua: $\mathcal{\zeta}= 0.69$, \textit{Faith}= $97.8\pm2.0\%$, Similarity = 0.60.}}\\
    \end{subfigure}
\vspace*{1mm} \hrule \vspace*{1mm}
    \begin{subfigure}[tc]{0.18\linewidth}
        \raisebox{-\height}{\includegraphics[width=\linewidth]{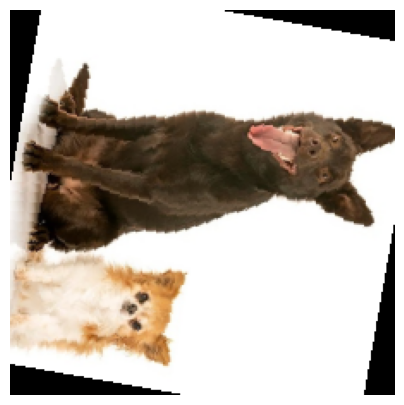}} \vspace*{-1mm}\\
        \centerline{\footnotesize ($\tau = 180\degree$)}
    \end{subfigure}
    \begin{subfigure}[c]{0.8\linewidth}
        \centering
        \raisebox{-\height}{\includegraphics[width=0.165\linewidth]{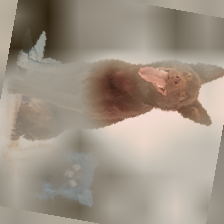}}\hspace*{0.5mm} \raisebox{-\height}{\includegraphics[width=0.82\linewidth]{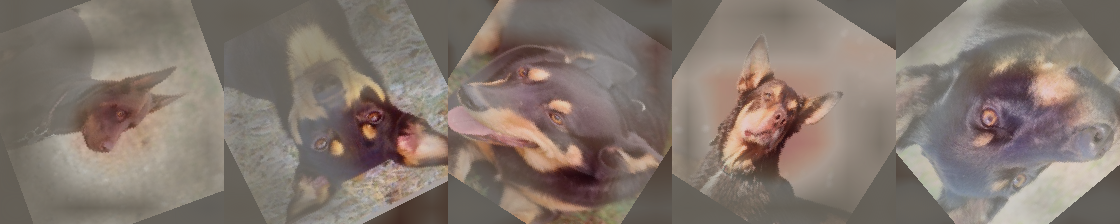}}\hspace{0.25\linewidth}\vspace*{-4mm}\\
        \raisebox{-\height}{\leftline{\footnotesize Kelpie: $\mathcal{\zeta}= 0.70$, \textit{Faith}= $97.6\pm2.2\%$, Similarity = 0.62.}}\\\vspace*{-1.2mm}
        \raisebox{-\height}{\includegraphics[width=0.165\linewidth]{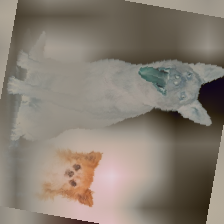}}\hspace*{0.5mm} \raisebox{-\height}{\includegraphics[width=0.82\linewidth]{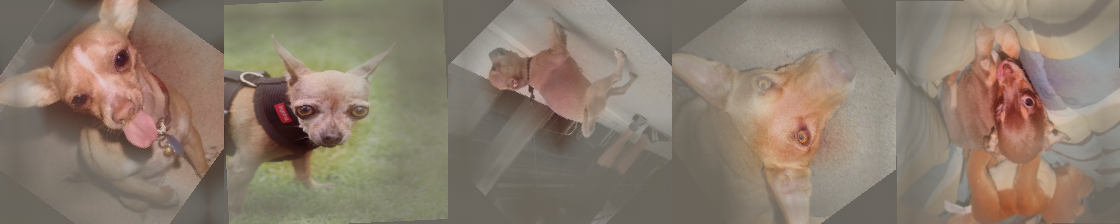}}\vspace*{0.7mm}\\
        \raisebox{-\height}{\leftline{\footnotesize Chihuahua: $\mathcal{\zeta}= 0.72$, \textit{Faith}= $95.4 \pm 4.2\%$, Similarity = 0.61.}}\\
    \end{subfigure}
    \vspace*{1mm} \hrule \vspace*{3mm}
    \hfill
\centerline{(a)}\medskip\vspace*{2mm}

    \begin{minipage}[b]{.49\linewidth}
      \centerline{\includegraphics[width=1\linewidth]{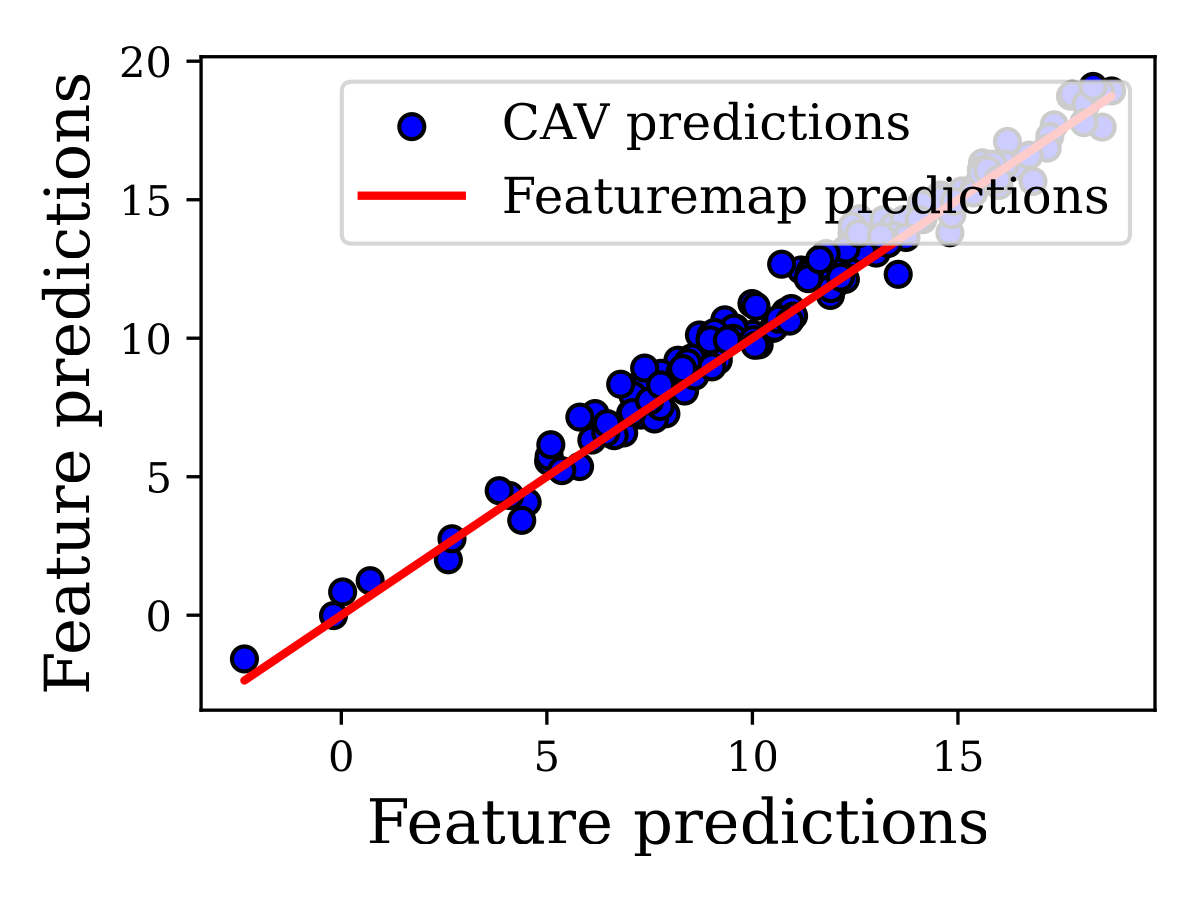}}
      \vspace*{-12ex}  \hspace*{2ex}
        \begin{center}
       {\hspace*{8ex} \small Kelpie: $\tau = 45\degree$  }
        \end{center}\vspace*{3mm}\medskip
        \vspace*{12ex} \hspace*{.2ex} 
    \end{minipage} 
    \begin{minipage}[b]{.49\linewidth}
      \centerline{\includegraphics[width=1\linewidth]{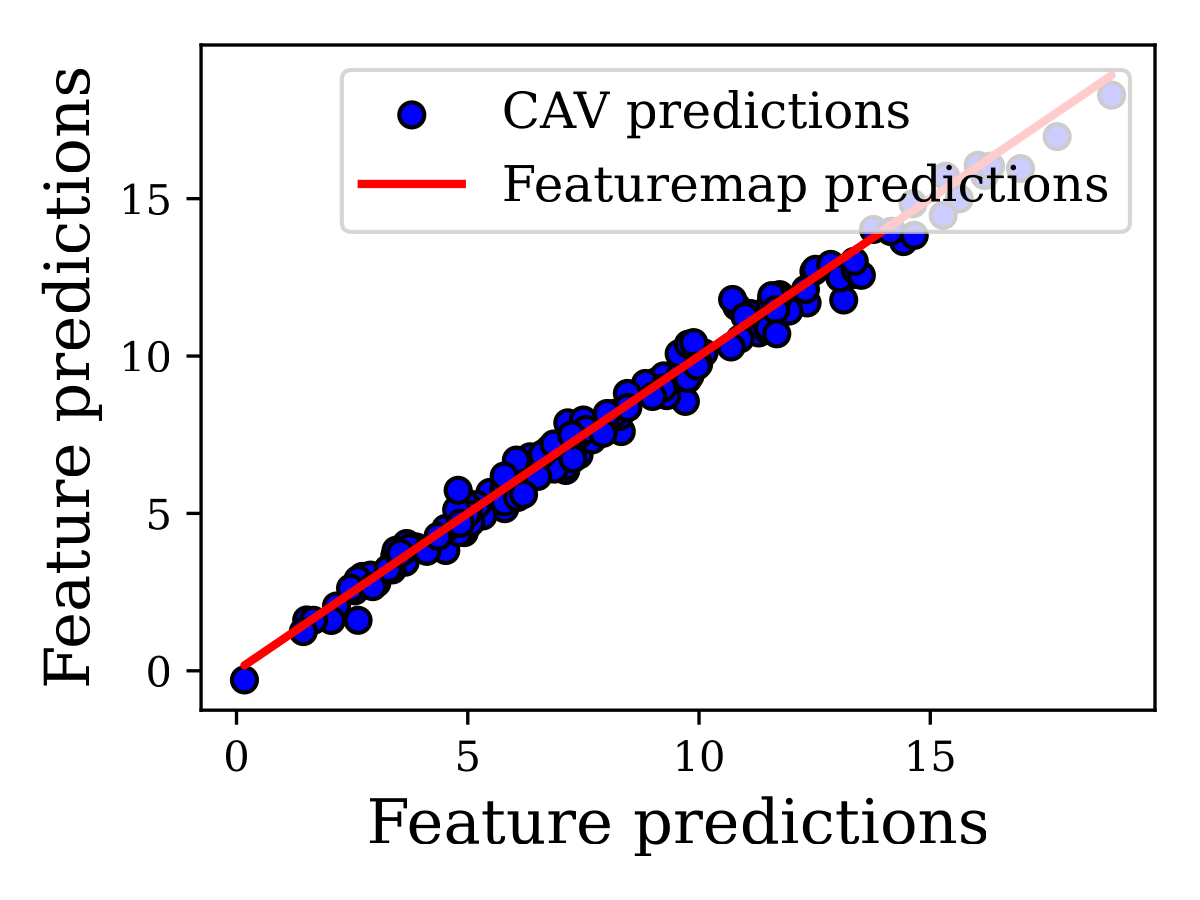}}
      \vspace*{-12ex} \hspace*{-.2ex}
        \begin{center}
        {\hspace*{7ex} \small Kelpie: $\tau = 180\degree$  }
        \end{center}\vspace*{3mm} \medskip
        \vspace*{12ex} \hspace*{.2ex} 
    \end{minipage}\vspace*{-22mm}\vspace*{1mm}
\centerline{(b)}
\caption{Top ResNet50 explanations for an Australian Kelpie and a Chihuahua (a) with image rotation ($\tau = \{45\degree, 90\degree, 180\degree\}$) and (b) faithfulness at $\tau = 45\degree$ and $180\degree$.}
\label{rotate}
\end{figure}

Rarely, discovered concepts did not meet the criteria for meaningfulness. To reflect this, we conducted sanity checks on TraNCE by introducing adversarial attacks on the images using different image transformations. We studied the impact of additive noise, rotation, and warping on the quality of concepts discovered and the degree of faithfulness of TraNCE. The results of these sanity checks on TraNCE are presented in Figs.~\ref{noisy}, ~\ref{warp}, and ~\ref{rotate} for ResNET50 and InceptionV3 using the test images presented in Fig. \ref{sys_model}. In our study, all the image transformations were done during TraNCE training/inference at $c^\prime$ = 32 with the same VAE settings. The Gaussian noise, $\varphi (\mu, \sigma)$ was added at zero mean ($\mu$) and varying standard deviation ($\sigma = \{0.1, 0.3, 0.5\}$), image warping was done with a constant value of smoothness ($\lambda = 50$) at varying displacement values ($\alpha = \{50.0, 100.0, 200.0\}$), and image rotations were performed at varying degrees ($\tau = \{45\degree, 90\degree, 180\degree$\}). 

The study reveals that higher image transformations are likely to decrease classifier accuracy \cite{DLNOISYLABELS}. This interference/distortion results in inaccurate, inconsistent, complex/ambiguous concepts with misplaced prototypes, reflected in the similarity scores, $\zeta$ values and \textit{Faith} scores. This invariably exposes inaccurate, inconsistent, and complex concepts with decreasing $\zeta$, \textit{Faith} and Similarity scores at increased additive noise levels as shown in Fig. \ref{noisy}(a). Warping noticeably impairs the explainer's performance on the test images, akin to introducing Gaussian noise, as evidenced in Fig. \ref{warp}. As shown in the marginally decreasing \textit{Faith} values in Fig. \ref{warp}(a), warping marginally affects explainer faithfulness (less than 5\%) but can lead to the discovery of ambiguous concept explanations and misplaced prototypes for a target class. This is evidenced in the reduced similarity scores and $\zeta$ values as $\alpha$ increases. On the other hand, the \textit{Faith} scores at different image rotation angles reveal a minimal change in explainer faithfulness and CNN accuracy. These are reflected in the stable similarity scores, $\zeta$ values, \textit{Faith} scores at increased $\tau$ values, and faithfulness visualization results in Fig. \ref{rotate}(b). In contrast, Fig. \ref{noisy}(b) and \ref{warp}(b) further show that increasing $\sigma$ and $\alpha$ values increase the sparsity in CAV predictions (blue dots) compared to the feature map predictions (red line). Overall, image transformation affects classifier accuracy due to many factors, including the type and extent of transformation applied, the dataset characteristics, and any accompanying preprocessing techniques, if any \cite{image_rotation}. These findings demonstrate that TraNCE passes the sanity check, reflecting CNN's logic.

\subsection{A deeper look into TraNCE}

\begin{figure}[!t]
\centering
\begin{subfigure}[tc]{0.18\linewidth}
\centerline{query image}\vspace*{-1mm}
        \raisebox{-\height}{\includegraphics[width=\linewidth]{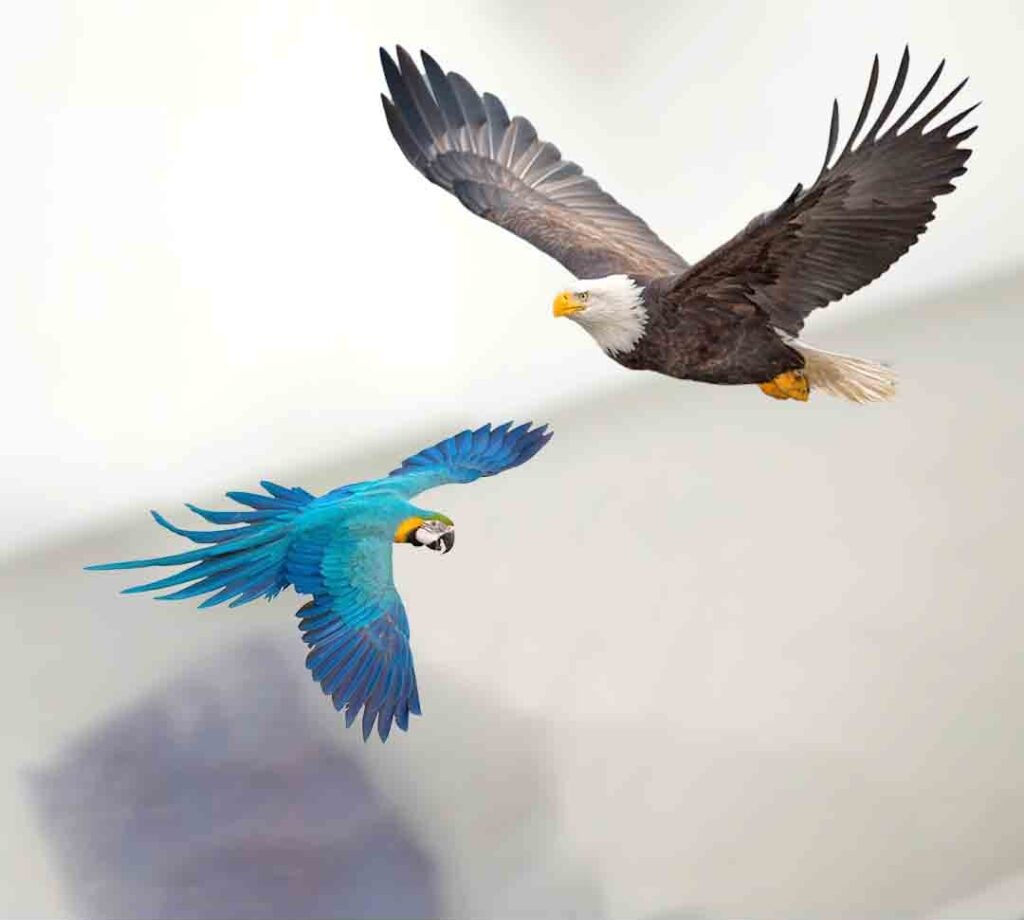}} \vspace*{-1mm}
    \end{subfigure}
    \begin{subfigure}[c]{1\linewidth}
        \centering 
        \hspace*{-1mm}\leftline{ \textbf{Concept} }\vspace*{-4mm} \hspace*{-68mm}\rightline {\textbf{Prototypes}}\vspace*{1mm}\\
        \hrule \vspace*{1mm}
        
        \raisebox{-\height}{\includegraphics[width=0.165\linewidth]{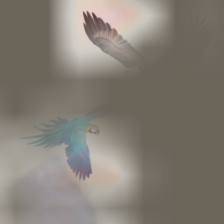}}\hspace*{0.5mm} \raisebox{-\height}{\includegraphics[width=0.82\linewidth]{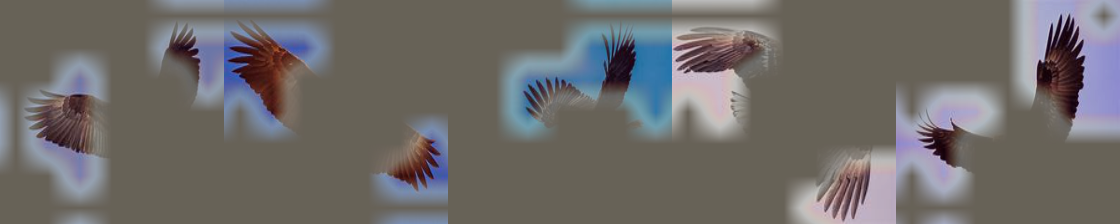}}\hspace{0.25\linewidth}\\\vspace*{0.7mm}
        \raisebox{-\height}{\centerline{Eagle: $\mathcal{\zeta}= 0.82$, \textit{Faith}= $98.6\pm1.4\%$, Similarity = 0.71.}}\\
        \raisebox{-\height}{\includegraphics[width=0.165\linewidth]{deeper/feature_29.png}}\hspace*{0.5mm} \raisebox{-\height}{\includegraphics[width=0.82\linewidth]{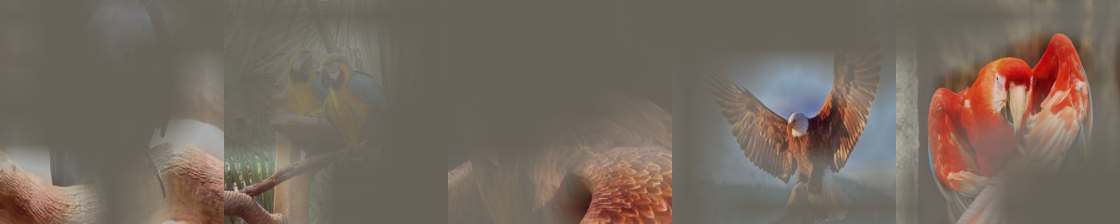}}\hspace{0.25\linewidth}\\\vspace*{0.7mm}
        \raisebox{-\height}{\centerline{Macaw: $\mathcal{\zeta}= 0.82$, \textit{Faith}= $98.6\pm1.4\%$, Similarity = 0.71.}}\\
        \vspace*{2mm}\hrule
        
        \vspace*{3mm}\centerline{(a)}\vspace*{3mm}

        \hrule \vspace*{1mm}
        \raisebox{-\height}{\includegraphics[width=0.165\linewidth]{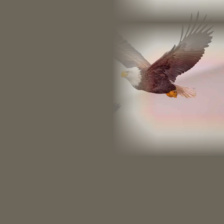}}\hspace*{0.5mm} \raisebox{-\height}{\includegraphics[width=0.82\linewidth]{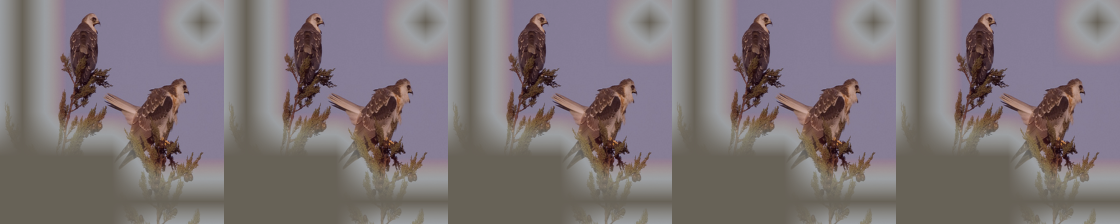}}\hspace{0.25\linewidth}\\\vspace*{0.7mm}
        \raisebox{-\height}{\centerline{Eagle: $\mathcal{\zeta}= 0.82$, \textit{Faith}= $98.6\pm1.4\%$, Similarity = 0.71.}}\\\vspace*{0.5mm}
        \raisebox{-\height}{\includegraphics[width=0.165\linewidth]{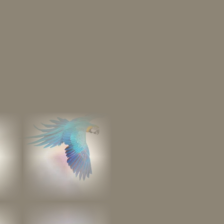}}\hspace*{0.5mm} \raisebox{-\height}{\includegraphics[width=0.82\linewidth]{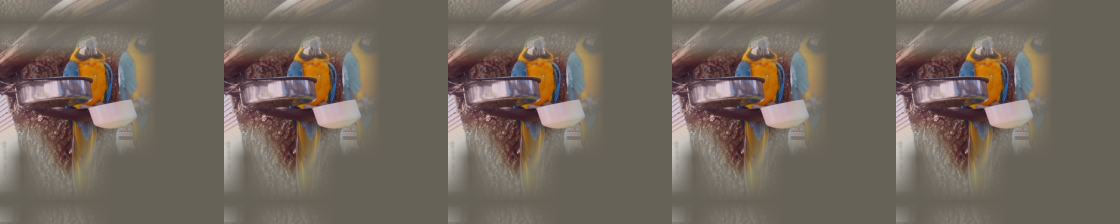}}\hspace{0.25\linewidth}\\\vspace*{0.7mm}
        \raisebox{-\height}{\centerline{Macaw: $\mathcal{\zeta}= 0.82$, \textit{Faith}= $98.6\pm1.4\%$, Similarity = 0.71.}}\\
        \vspace*{2mm}\hrule

        \vspace*{3mm}\centerline{(b)}\vspace*{3mm}

        \hrule \vspace*{1mm}
        \raisebox{-\height}{\includegraphics[width=0.165\linewidth]{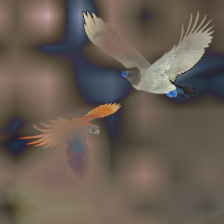}}\hspace*{0.5mm} \raisebox{-\height}{\includegraphics[width=0.82\linewidth]{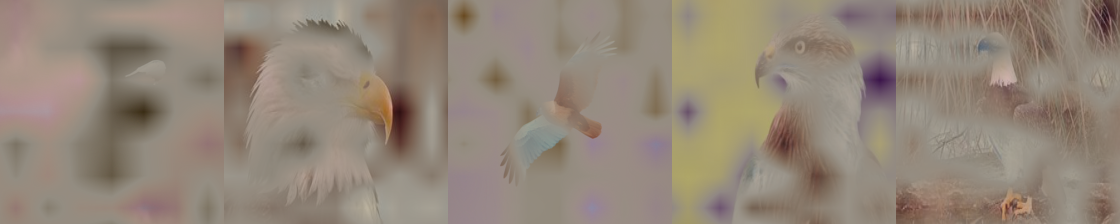}}\hspace{0.25\linewidth}\\\vspace*{0.7mm}
        \raisebox{-\height}{\centerline{Eagle: $\mathcal{\zeta}= 0.82$, \textit{Faith}= $98.6\pm1.4\%$, Similarity = 0.71.}}\\\vspace*{0.5mm}
        \raisebox{-\height}{\includegraphics[width=0.165\linewidth]{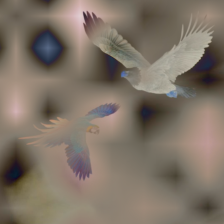}}\hspace*{0.5mm} \raisebox{-\height}{\includegraphics[width=0.82\linewidth]{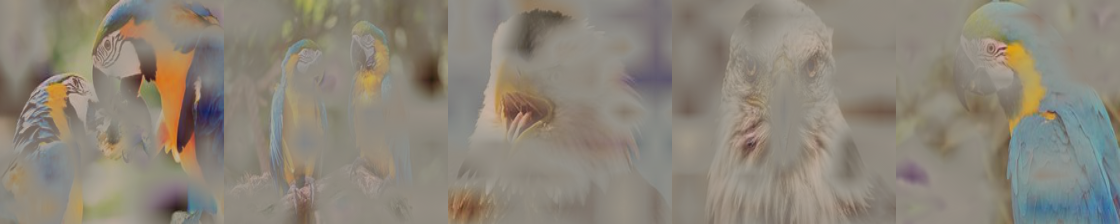}}\hspace{0.25\linewidth}\\\vspace*{0.7mm}
        \raisebox{-\height}{\centerline{Macaw: $\mathcal{\zeta}= 0.82$, \textit{Faith}= $98.6\pm1.4\%$, Similarity = 0.71.}}\\
        \vspace*{2mm}\hrule
        \vspace*{2mm}\centerline{(c)}\vspace*{2mm}
    \end{subfigure}
\caption{InceptionV3 explanations for an Eagle and a Macaw at $c^\prime = 32$, revealing (a) concept explanations for two classes with a high inter-class similarity (b) duplicate concept explanations, and (c) ambiguous concept explanations.}
\label{interclass}
\end{figure}

Amid the efficiencies of TraNCE and its sanity for discovering concept explanations consistent with human-understandable features/attributes for a target class, a deeper look into its performance reveals a few limitations and potentials. One of these limitations is the discovery of global concepts for target classes with high inter-class similarity. For instance, the discovery of the concept ``feathers" for the Eagle and Macaw, as shown in Fig. \ref{interclass} (a). As shown, TraNCE, like other concept-based explainers generates not just the concept feathers but also prototypes sourced from both classes' folders in the dataset. This presents the question \textit{whose feathers were discovered?}

Rarely, TraNCE may also discover duplicate prototypes as shown in Fig.~\ref{interclass} (b) and ambiguous prototypes as shown in Fig.~\ref{interclass} (c), raising more questions about \textit{how does a CNN perceive similar concepts differently?} Despite these concerns, such rare complex/ambiguous concepts may align with CNN logic such that they may make sense to CNN but are not human-understandable. A potential solution would be to improve the discriminatory power of the CNN to accurately distinguish between classes with high inter-class similarity as recent works \cite{zicheng,lie,exactly} show so that the explainer may better reflect the CNN logic. On the other hand, the explainer may also be equipped with contrasting paradigms in their architecture to prioritize contrasting concept explanations that improve explainer faithfulness for pre-trained CNNs with limited classification accuracies. We attempted this by leveraging the standard deviation of the $\zeta$ values of discovered concepts for local explanations and choosing a threshold value of 0.5, above which discovered concepts are highly contrasting. We present our findings in Table~\ref{table_contrast}. For each CNN classifier in Table~\ref{table_contrast}, we set a maximum $c^\prime = 32$ and retrieved the number of contrasting concepts $c^\prime_{\delta}$ for each class. We also recorded the \textit{Faith} scores for each case. Consistent with our findings for the dog and bird classes, Table~\ref{table_contrast} shows that although $c^\prime$ is 32, only 15 to 21 concepts are highly contrasting for each CNN. At the same time, the rest are duplicated, ambiguous and/or complex without affecting the \textit{Faith} scores. While this is still in its early stages, our preliminary findings reveal that integrating a contrasting paradigm could enhance the quality of concept explanations. Future research could investigate this further. 

\begin{table*}[!tbph]
 \centering
\caption{Number of contrasting concepts discovered by TraNCE for each CNN at $c^\prime$ = 32 ($\delta = 0.5$).}
\label{table_contrast}
\begin{tabular}{P{60pt}P{4pt}P{44pt}|P{4pt}P{44pt}|P{4pt}P{44pt}|P{4pt}P{44pt}|P{4pt}P{44pt}|P{4pt}P{44pt}}
    \hline
     &\multicolumn{2}{c}{ResNet18 \cite{resnet}} & \multicolumn{2}{c}{ResNet50 \cite{resnet}} & \multicolumn{2}{c}{ResNext101 \cite{resnext}} & \multicolumn{2}{c}{WideResNext50 \cite{wideresnet}} & \multicolumn{2}{c}{WideResNext101 \cite{wideresnet}} & \multicolumn{2}{c}{InceptionV3 \cite{inceptionv3}}\\
    \hline
    Target Class & $c^\prime_{\delta}$ & \textit{Faith} & $c^\prime_{\delta}$ & \textit{Faith}& $c^\prime_{\delta}$ & \textit{Faith}& $c^\prime_{\delta}$ & \textit{Faith}& $c^\prime_{\delta}$ & \textit{Faith}& $c^\prime_{\delta}$ & \textit{Faith}\\
    \hline
  Australian Kelpie  & 21  & $97.9 \pm 2.2\%$ & 17 & $98.2 \pm 1.1\%$ & 22 & $96.3 \pm 2.4\%$ & 21  & $97.2 \pm 1.3\%$ & 19 & $95.8 \pm 3.1\%$ & 18 &  $97.9 \pm 2.2\%$\\ 
   Chihuahua  & 19  & $94.9 \pm 3.4\%$ & 20 & $98.8 \pm 1.0\%$ & 22 & $93.7 \pm 5.1\%$ & 22  &  $96.2 \pm 2.2\%$ & 21 & $89.8 \pm 5.9\%$ & 21 & $99.0 \pm 0.6\%$\\
  Eagle & 19 & $93.7 \pm 5.8\%$ & 18 & $98.7 \pm 1.2\%$ & 21 & $95.5 \pm 3.6\%$ & 23  & $96.2 \pm 2.9\%$ & 22 & $94.2 \pm 4.8\%$ & 18 & $98.6 \pm 1.2\%$\\
  Macaw &  21 & $97.1 \pm 2.2\%$ & 21 & $98.4 \pm 1.1\%$ & 22 & $92.2 \pm 5.6\%$ & 20  & $97.1 \pm 1.7\%$ & 21 & $92.7 \pm 6.9\%$ & 18 & $98.8 \pm 1.1\%$\\
    Tiger Cat &  20 & $97.1 \pm 2.2\%$ & 21 & $98.4 \pm 1.1\%$ & 22 & $92.2 \pm 5.6\%$ & 20  & $97.1 \pm 1.7\%$ & 21 & $92.7 \pm 6.9\%$ & 18 & $98.8 \pm 1.1\%$\\

    Birdhouse &  18 & $96.5 \pm 3.2\%$ & 19 & $92.4 \pm 7.2\%$ & 17 & $92.9 \pm 4.6\%$ & 16  & $97.3 \pm 1.0\%$ & 18 & $94.0 \pm 2.9\%$ & 18 & $91.7 \pm 5.2\%$\\

    Indigo Buting &  24 & 9$0.8 \pm 6.3\%$ & 21 &$ 93.4 \pm 4.6\%$ & 19 & $94.7 \pm 2.6\%$ & 17  & $96.8 \pm 2.4\%$ & 20 & $94.2 \pm 3.0\%$ & 18 & $93.1 \pm 2.7\%$\\
    Cassette &  17 & $97.1 \pm 2.2\%$ & 15 & $97.5 \pm 0.5\%$ & 17 & $97.7 \pm 1.6\%$ & 16  & $91.9 \pm 5.7\%$ & 20 & $91.4 \pm 5.7\%$ & 19 & $92.8 \pm 4.2\%$\\

    Sunglasses &  17 & $93.6 \pm 2.9\%$ & 18 & $97.5 \pm 1.5\%$ & 16 & $95.2 \pm 2.6\%$ & 20  & $94.1 \pm 2.7\% $& 15 & $92.3 \pm 4.9\%$ & 18 & $91.2 \pm 5.1\%$\\
  
  \hline
\end{tabular}
\end{table*}

\subsection{A Broader Perspective}
\label{sec:Perspective}

The proposed TraNCE framework, along with other concept-based explainability frameworks \cite{tcav, ice, ace, craft}, contributes to making AI more human-centric. Despite being in its early stages, concept-based explanations offer straightforward insights into what CNNs see in images. The unique \textit{Faith} component in TraNCE represents a step toward enhancing the comprehensibility of CNN explanations, moving beyond merely relying on the Fidelity of identified concepts on the one hand and a step towards relying less on humans for assessing the concepts for consistency. Integrating Coherence with Fidelity in the \textit{Faith} metric offers deeper quantitative validation of their meaningfulness as reflected in the respective \textit{Faith} scores, suggesting that concepts that are imperfect, ambiguous or complex with consistent prototypes (low Fidelity, high Coherence) can be accepted but should be rejected if the prototype explanations are inconsistent (low Coherence), amidst high Fidelity. In sharp contrast, concept explanations with high Fidelity and Coherence should be accepted while concepts with low Fidelity and Coherence scores should be rejected. In summary, concept and prototype consistency, even with low Fidelity scores, enhances trust in the generated explanations. Future work will address the identified limitations.

Given that the feature maps generated through the convolution and filtering operations of CNN models are continuous, and the VAE's efficiency with continuous data \cite{vae_continuous}, we see significant promise in the VAE's potential in the domain, particularly due to its adaptable architecture. While computational cost may be a concern, we believe the computational cost is an acceptable trade-off, given the potential risks associated with poor explainer faithfulness provided by simpler reducers like the NMF and PCA. Moreover, the comparatively higher computational cost is not reflected in the inference time. Our proposed TraNCE framework provides the avenue of saving a trained explainer for future inference without a need to re-train, offering transferability and adaptation. Furthermore, selecting prototypical parts doesn't necessitate storing raw training data, a step toward the General Data Protection Regulation (GDPR) compliance by reducing the amount of personal data retained and minimizing the risk of data breaches \cite{gdpr}. However, VAEs are sensitive to the choice of hyperparameters, especially in the dimensionality of the latent space \cite{vae_parameters, vae_parameters2}. This would imply that a poor choice of hyperparameters may lead to sub-optimal reduction performance and require expert knowledge for optimal utility. Additionally, there may be a need to establish a generalization paradigm for VAE to optimize TraNCE efficiency for FGVC tasks with varying inter-class variances. Nonetheless, we believe that exploring adaptable explainers could be expected in the near future. 

From a comprehensive point of view, relying only on visual concepts for discrimination may not be optimal because certain discriminating features are not visually recognizable e.g. behaviours, demeanour, temperaments, etc. Typical cases are the discrimination between two identical twins or emotion recognition. Such limitations necessitate video categorization and expanding TraNCE's capabilities for video classifiers. However, since human-centred explanations typically focus on identifying the most crucial concepts rather than attempting to uncover every conceivable concept (like concept-based explainers do), it is advisable to prioritize a select few of the most critical concepts to minimize complexity. A future line of work should prioritize the most important (contrasting) concepts, as we have attempted in Table~\ref{table_contrast} or combine duplicate concepts into one representative concept. Filling this gap would require an interdisciplinary approach involving computer science, cognitive science, and neuroscience~\cite {visual_answer, review_xai}. From a different perspective, the proposed TraNCE in its current form effectively elucidates CNNs' sequential data processing and local feature capture. The concept visualization, as we have demonstrated, provides qualitative explanations (where the CNN looked) which although prone to bias, each prototypical representation further improves trust because the prototypes validate what was seen, not just where was looked. These in addition to the quantitative checkpoints provide an even stronger avenue for ensuring trust. Notwithstanding, the proposed TraNCE's suitability for transformers, which operate differently by capturing global dependencies among input tokens in parallel, remains uncertain. Exploring this path could be equally challenging as a research direction.

\section{Conclusion}
\label{conclusion}
In this work, we propose TraNCE, a concept-based explanation method for CNNs. TraNCE autonomously discovers meaningful and consistent high-level concepts as prototypical parts of a query image to provide human-understandable explanations. To enhance the interpretability and reliability of our method, TraNCE incorporates a novel strategy for minimizing decomposition losses. This involves the application of a non-linear dimensionality reduction technique, ensuring that the identified concepts accurately capture the essential components of the input image. Furthermore, we introduce a quantitative measure termed the \textit{Faith} score, aimed at evaluating the consistency of these discovered concepts comprehensively with the concept Fidelity. This metric provides a more reliable assessment of the faithfulness of our explanation method, enhancing its overall effectiveness.

Our experiments, conducted on FGVC tasks, demonstrate the consistent capability of TraNCE to unveil the decision logic embedded within a CNN. By elucidating \textit{what} the CNN saw, our method offers valuable insights into the underlying processes of the neural network. It reflects its performance under adversarial attacks by providing both qualitative and quantitative insights. Our study also highlights challenges in the concept-based explainability domain and discusses recommendations for future research directions. These contributions not only establish benchmark competencies in the field but also foster further advancements.

\bibliographystyle{IEEEtran}
\bibliography{refs}

\vfill

\end{document}